\documentclass{article}

\usepackage[final]{neurips_2019}

\usepackage[utf8]{inputenc} 
\usepackage[T1]{fontenc}    
\usepackage{url}            
\usepackage{booktabs}       
\usepackage{amsfonts}       
\usepackage{nicefrac}       
\usepackage{microtype}      
\usepackage{enumitem}

\usepackage{color,xcolor}
\usepackage{epsfig}
\usepackage{graphicx}

\usepackage{adjustbox}
\usepackage{array}
\usepackage{booktabs}
\usepackage{colortbl}
\usepackage{float,wrapfig}
\usepackage{hhline}
\usepackage{multirow}
\usepackage{subcaption} 

\usepackage{amsmath,amsfonts,amsthm,amssymb}
\usepackage{bm}
\usepackage{nicefrac}
\usepackage{microtype}

\usepackage{changepage}
\usepackage{extramarks}
\usepackage{fancyhdr}
\usepackage{lastpage}
\usepackage{setspace}
\usepackage{soul}
\usepackage{xspace}

\usepackage{url}

\usepackage{algorithmic}
\usepackage{algorithm}
\usepackage{enumerate}
\usepackage{todonotes} 

\usepackage{titlesec}
\usepackage{makecell}

\usepackage{amsmath,amsthm,amssymb}
\usepackage{mathtools}

\newcolumntype{L}[1]{>{\raggedright\let\newline\\\arraybackslash\hspace{0pt}}m{#1}}
\newcolumntype{C}[1]{>{\centering\let\newline\\\arraybackslash\hspace{0pt}}m{#1}}
\newcolumntype{R}[1]{>{\raggedleft\let\newline\\\arraybackslash\hspace{0pt}}m{#1}}

\newcommand{\eqn}[1]{Equation~\ref{#1}}
\newcommand{\fig}[1]{Figure~\ref{#1}}
\newcommand{\tbl}[1]{Table~\ref{#1}}
\newcommand{\alg}[1]{Algorithm~\ref{#1}}

\newcommand{\ignore}[1]{}

\DeclareMathOperator*{\argmin}{arg\,min}

\makeatletter
\DeclareRobustCommand\onedot{\futurelet\@let@token\@onedot}
\def\@onedot{\ifx\@let@token.\else.\null\fi\xspace}

\makeatother

\definecolor{MyDarkBlue}{rgb}{0,0.08,1}
\definecolor{MyDarkGreen}{rgb}{0.02,0.6,0.02}
\definecolor{MyDarkRed}{rgb}{0.8,0.02,0.02}
\definecolor{MyDarkOrange}{rgb}{0.40,0.2,0.02}
\definecolor{MyPurple}{RGB}{111,0,255}
\definecolor{MyRed}{rgb}{1.0,0.0,0.0}
\definecolor{MyGold}{rgb}{0.75,0.6,0.12}
\definecolor{MyDarkgray}{rgb}{0.66, 0.66, 0.66}

\newcommand{\myparagraph}[1]{\vspace{-10pt}\paragraph{#1}}
\titlespacing*{\section}{0pt}{0pt plus 2pt minus 2pt}{0pt plus 2pt minus 2pt}
\titlespacing\subsection{0pt}{0pt plus 1pt minus 1pt}{0pt plus 1pt minus 1pt}

\DeclarePairedDelimiterX{\infdivx}[2]{(}{)}{%
  #1\;\delimsize|\delimsize|\;#2%
}

\title{Implicit Generation and Modeling with Energy-Based Models}

\author{
  Yilun Du \thanks{Work done at OpenAI}\\
  MIT CSAIL\\
  \And
  Igor Mordatch \\
  Google Brain\\
  \And
}

\begin{document}

\maketitle
\vskip 0.3in

\begin{abstract}
Energy based models (EBMs) are appealing due to their generality and simplicity in likelihood modeling, but have been traditionally difficult to train. We present techniques to scale MCMC based EBM training on continuous neural networks, and we show its success on the high-dimensional data domains of ImageNet32x32, ImageNet128x128, CIFAR-10, and robotic hand trajectories, achieving better samples than other likelihood models and nearing the performance of contemporary GAN approaches, while covering all modes of the data. We highlight some unique capabilities of implicit generation such as compositionality and corrupt image reconstruction and inpainting. Finally, we show that EBMs are useful models across a wide variety of tasks, achieving state-of-the-art out-of-distribution classification, adversarially robust classification, state-of-the-art continual online class learning, and coherent long term predicted trajectory rollouts.  \footnotetext{Correspondence to: yilundu@mit.edu} \footnotetext{Additional results, source code, and pre-trained models are available at https://sites.google.com/view/igebm}

\end{abstract}

\section{Introduction}



Learning models of the data distribution and generating samples are important problems in machine learning for which a number of methods have been proposed, such as Variational Autoencoders (VAEs) \citep{Kingma2014Auto} and Generative Adversarial Networks (GANs) \citep{Goodfellow2014Generative}.In this work, we advocate for continuous energy-based models (EBMs), represented as neu
ral networks, for generative modeling tasks and as a building block for a wide variety of tasks. These models aim to learn an energy function $E(\mathbf{x})$ that assigns low energy values to inputs $\mathbf{x}$ in the data distribution and high energy values to other inputs. Importantly, they allow the use of an \emph{implicit} sample
 generation procedure, where sample $\mathbf{x}$ is found from $\mathbf{x} \sim e^{-E(\mathbf{x})}$ through MCMC sampling.
Combining implicit sampling with energy-based models for generative modeling has a number of conceptual advantages compared to methods such as VAEs and GANs which use explicit functions to generate samples:

\textbf{Simplicity and Stability:} An EBM is the only object that needs to be trained and designed. Separate networks are not tuned to ensure balance (for example, \citep{he2019lagging} point out unbalanced training can result in posterior collapse in VAEs or poor performance in GANs \citep{kurach2018gan}). 

\textbf{Sharing of Statistical Strength:} Since the EBM is the only trained object, it requires fewer model parameters than approaches that use multiple networks. More importantly, the model being concentrated in a single network allows the training process to develop a shared set of features as opposed to developing them redundantly in separate networks. 

\textbf{Adaptive Computation Time:} Implicit sample generation in our work is an iterative stochastic optimization process, which allows for a trade-off between generation quality and computation time. This allows for a system that can make fast coarse guesses or more deliberate inferences by running the optimization process longer. It also allows for refinement of external guesses.

\textbf{Flexibility Of Generation:} The power of an explicit generator network can become a bottleneck on the generation quality. For example, VAEs and flow-based models are bound by the manifold structure of the prior distribution and consequently have issues modeling discontinuous data manifolds, often assigning probability mass to areas unwarranted by the data. EBMs avoid this issue by directly modeling particular regions as high or lower energy. 

\textbf{Compositionality:} If we think of energy functions as costs for a certain goals or constraints, summation of two or more energies corresponds to satisfying all their goals or constraints \citep{mnih2004learning,haarnoja2017reinforcement}. While such composition is simple for energy functions (or product of experts \citep{hinton1999products}), it induces complex changes to the generator that may be difficult to capture with explicit generator networks. 

\begin{wrapfigure}{l}{0.5\textwidth}
\begin{center}
    \includegraphics[width=1.0\linewidth]{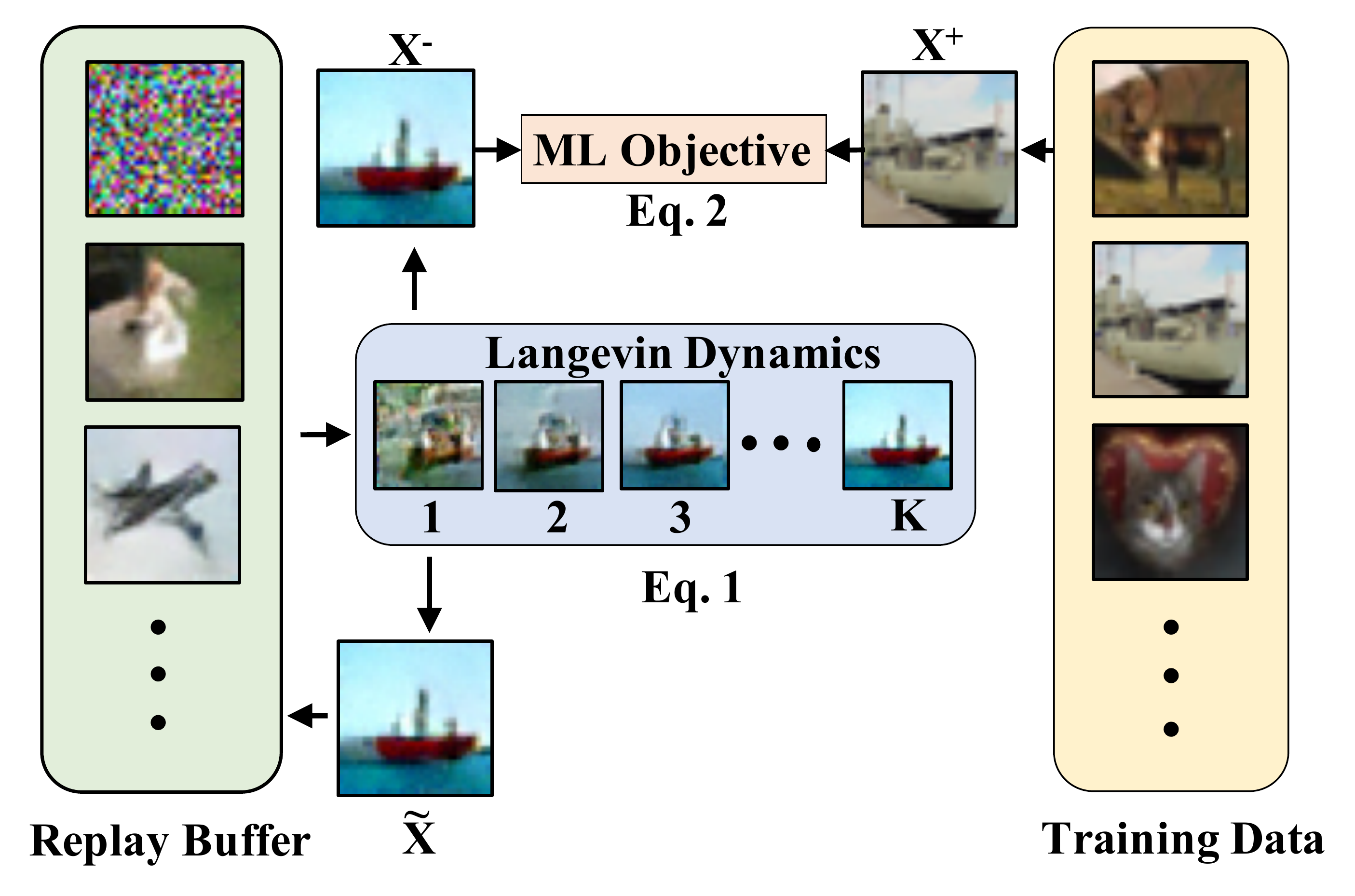}
\end{center}

\caption{\small Overview of our method and the interrelationship of the components involved.}  
\label{fig:overview}
\end{wrapfigure}

Despite these advantages, energy-based models with implicit generation have been difficult to use on complex high-dimensional data domains. 
 In this work, we use Langevin dynamics \citep{welling2011bayesian}, which uses gradient information for effective sampling and initializes chains from random noise for more mixing. We further maintain a replay buffer of past samples (similarly to \citep{tieleman2008training} or \citep{mnih2013playing}) and use them to initialize Langevin dynamics to allow mixing between chains.
 An overview of our approach is presented in \fig{fig:overview}.

Empirically, we show that energy-based models trained on CIFAR-10 or ImageNet image datasets generate higher quality image samples than likelihood models and nearing that of contemporary GANs approaches, while not suffering from mode collapse. The models exhibit properties such as correctly assigning lower likelihood to out-of-distribution images than other methods (no spurious modes) and generating diverse plausible image completions (covering all data modes). Implicit generation allows our models to naturally denoise or inpaint corrupted images, convert general images to an image from a specific class, and generate samples that are compositions of multiple independent models.

Our contributions in this work are threefold. Firstly, we present an algorithm and techniques for training energy-based models that scale to challenging high-dimensional domains.  Secondly, we highlight unique properties of energy-based models with implicit generation, such as compositionality and automatic decorruption and inpainting. Finally, we show that energy-based models are useful across a series of domains, on tasks such as out-of-distribution generalization, adversarially robust classification, multi-step trajectory prediction and online learning. 
\section{Related Work}

Modeling data using the Boltzmann distribution has been used extensively across diverse fields. Such models include Ising models \citep{cipra1987introduction} in electromagnetism, Markov Logic Networks \citep{richardson2006markov} over knowledge bases, the Helmholtz \citep{Dayan1995helmholtz} and Boltzmann Machines \citep{Ackley1985learning} in machine learning, and the FRAME model \citep{zhu1998filters} in computer vision. 

In the computer vision community, the FRAME model \citep{zhu1998filters, wu2000equivalence} utilizes the Boltzmann distribution or the Gibbs distribution to represent different textures of images, as well as a model of texture perception.  The FRAME model is further extended for modeling object patterns in \citep{xie2015learning, xie2016inducing} \citet{lu2016learning} uses deep network features in the FRAME model for image synthesis. \citet{xie2016theory}  further extends the FRAME model to the energy-based generative ConvNet, in which the energy function is parameterized by a ConvNet structure. A multi-grid sampling and training strategy for the energy-based generative ConvNet is also studied in \citep{gao2018learning}.

In the deep learning community, such models are are known as Energy-based models (EBMs). \cite{Ackley1985learning,hinton2006training,salakhutdinov2009deep} proposed latent based EBMs where energy is represented as a composition of latent and observable variables. In contrast \cite{mnih2004learning, hinton2006unsupervised} proposed EBMs where inputs are directly mapped to outputs, a structure we follow. We refer readers to \citep{lecun2006tutorial} for a comprehensive tutorial on energy models. 

The primary difficulty in training EBMs comes from effectively estimating and sampling the partition function. One approach to train energy based models is to sample the partition function through amortized generation. \cite{kim2016deep,zhao2016energy,haarnoja2017reinforcement,kumar2019maximum} propose learning a separate network to generate samples, while \cite{xie2016cooperative, xie2018cooperative} use a separate network to initialize MCMC sampling,  which makes these methods closely connected to GANs \citep{finn2016connection}, but these methods do not have the advantages of implicit sampling noted in the introduction. Furthermore, amortized generation is prone to mode collapse, especially when training the sampling network without an entropy term which is often approximated or ignored. 

An alternative approach is to use MCMC sampling to directly estimate the partition function. This has an advantage of provable mode exploration and allows the benefits of implicit generation listed in the introduction. \citet{hinton2006training} proposed Contrastive Divergence, which uses gradient free MCMC chains initialized from training data to estimate the partition function. Similarly, \cite{salakhutdinov2009deep} apply contrastive divergence, while \cite{tieleman2008training} proposes PCD, which propagates MCMC chains throughout training. By contrast, we initialize chains from random noise, allowing each mode of the model to be visited with equal probability. But initialization from random noise comes at a cost of longer mixing times. As a result we use Gradient based MCMC (Langevin Dynamics) for more efficient sampling and to offset the increase of mixing time which was also studied previously in \citep{teh2003energy, xie2016theory}. We note that HMC \citep{Neal2011MCMC} may be an even more efficient gradient algorithm for MCMC sampling, though we found Langevin Dynamics to be more stable. To allow gradient based MCMC, we use continuous inputs, while most approaches have used discrete inputs. We build on idea of PCD and maintain a replay buffer of past samples to additionally reduce mixing times.

\def\x{\mathbf{x}}
\def\sx{\tilde{\x}}
\def\KL{\text{KL}}
\newcommand{\E}[2]{\mathbb{E}_{#1} \left[#2 \right] }
\newcommand{\ent}[1]{\mathcal{H}\left[ #1 \right]}

\section{Energy-Based Models and Sampling}
Given a datapoint $\x$, let $E_\theta(\x) \in \mathbb{R}$ be the energy function. In our work this function is represented by a deep neural network parameterized by weights $\theta$. The energy function defines a probability distribution via the Boltzmann distribution $p_\theta(\x) = \frac{\exp(-E_\theta(\x))}{Z(\theta)}$, where $Z(\theta)=\int \exp(-E_\theta(\x)) d\x$ denotes the partition function. Generating samples from this distribution is challenging, with previous work relying on MCMC methods such as random walk or Gibbs sampling \citep{hinton2006training}. These methods have long mixing times, especially for high-dimensional complex data such as images. To improve the mixing time of the sampling procedure, we use Langevin dynamics which makes use of the gradient of the energy function to undergo sampling
\begin{align}
\sx^k &= \sx^{k-1} - \frac{\lambda}{2} \nabla_\x E_\theta (\sx^{k–1}) + \omega^k, \; \omega^k \sim \mathcal{N}(0,\lambda)
\label{eq:langevin}
\end{align}
where we let the above iterative procedure define a distribution $q_\theta$ such that $\sx^K \sim q_\theta$. As shown by \cite{welling2011bayesian} as $K \rightarrow \infty$ and $\lambda \rightarrow 0$ then $q_\theta \rightarrow p_\theta$ and this procedure generates samples from the distribution defined by the energy function. Thus, samples are generated implicitly\footnote{Deterministic case of procedure in (\ref{eq:langevin}) is $\x = \argmin E(\x)$, which makes connection to implicit functions more clear.} by the energy function $E$ as opposed to being explicitly generated by a feedforward network.

In the domain of images, if the energy network has a convolutional architecture, energy gradient $\nabla_\x E$ in (\ref{eq:langevin}) conveniently has a deconvolutional architecture. Thus it mirrors a typical image generator network architecture, but without it needing to be explicitly designed or balanced. We take two views of the energy function $E$: firstly, it is an object that defines a probability distribution over data  and secondly it defines an implicit generator via (\ref{eq:langevin}). 

\subsection{Maximum Likelihood Training}
We want the distribution defined by $E$ to model the data distribution $p_D$, which we do by minimizing the negative log likelihood of the data $\mathcal{L}_\text{ML}(\theta) = \E{\x \sim p_D}{ -\log p_\theta(\x) }$ where
$-\log p_\theta(\x) = E_\theta(\x) - \log Z(\theta)$. This objective is known to have the gradient (see \citep{turner2005cd} for derivation) $\nabla_\theta \mathcal{L}_\text{ML} = \E{\x^+ \sim p_D}{\nabla_\theta E_\theta(\x^+)} - \E{\x^- \sim p_\theta}{\nabla_\theta E_\theta(\x^-)}$. Intuitively, this gradient decreases energy of the positive data samples $\x^+$, while increasing the energy of the negative samples $\x^-$ from the model $p_\theta$. We rely on Langevin dynamics in (\ref{eq:langevin}) to generate $q_\theta$ as an approximation of $p_\theta$:
\begin{align}
\label{eq:grad_likelihood}
\nabla_\theta \mathcal{L}_\text{ML} \approx \E{\x^+ \sim p_D}{\nabla_\theta E_\theta(\x^+)} - \E{\x^- \sim q_\theta}{\nabla_\theta E_\theta(\x^-)}.
\end{align}

This is similar to the gradient of the Wasserstein GAN objective \citep{arjovsky2017wasserstein}, but with an implicit MCMC generating procedure and no gradient through sampling. This lack of gradient is  important as it controls between the diversity in likelihood models and the mode collapse in GANs.  



The approximation in (\ref{eq:grad_likelihood}) is exact when Langevin dynamics generates samples from $p$, which happens after a sufficient number of steps (mixing time). We show in the supplement that $p_d$ and $q$ appear to match each other in distribution, showing evidence that $p$ matches $q$. We note that even in cases when a particular chain does not fully mix, since our initial proposal distribution is a uniform distribution, all modes are still equally likely to be explored.

\subsection{Sample Replay Buffer}
Langevin dynamics does not place restrictions on sample initialization $\sx^0$ given sufficient sampling steps. However initialization plays an crucial role in mixing time. Persistent Contrastive Divergence (PCD) \citep{tieleman2008training} maintains a single persistent chain to improve mixing and sample quality. We use a sample replay buffer $\mathcal{B}$ in which we store past generated samples $\sx$ and use either these samples or uniform noise to initialize Langevin dynamics procedure. This has the benefit of continuing to refine past samples, further increasing number of sampling steps $K$ as well as sample diversity. In all our experiments, we sample from $\mathcal{B}$ 95\% of the time and from uniform noise otherwise.

\subsection{Regularization and Algorithm}
Arbitrary energy models can have sharp changes in gradients that can make sampling with Langevin dynamics unstable. We found that constraining the Lipschitz constant of the energy network can ameliorate these issues. To constrain the Lipschitz constant, we follow the method of \citep{miyato2018spectral} and add spectral normalization to all layers of the model. Additionally, we found it useful to weakly L2 regularize energy magnitudes for both positive and negative samples during training, as otherwise while the difference between positive and negative samples was preserved, the actual values would fluctuate to numerically unstable values. Both forms of regularization also serve to ensure that partition function is integrable over the domain of the input, with spectral normalization ensuring smoothness and L2 coefficient bounding the magnitude of the unnormalized distribution. We present the algorithm below, where $\Omega(\cdot)$ indicates the stop gradient operator.

\begin{figure}[H]
\centering
\begin{minipage}{0.5\textwidth}
\begin{algorithm}[H]
\small
\begin{algorithmic}
    \STATE \textbf{Input:} data dist. $p_D(\x)$, step size $\lambda$, number of steps $K$
    \STATE $\mathcal{B} \gets \varnothing$
    \WHILE{not converged}
    \STATE $\x^+_i \sim p_D$
    \STATE $\x^0_i \sim \mathcal{B}$ with 95\% probability and $\mathcal{U}$ otherwise \vspace{2mm}
    \STATE \emph{$\triangleright$ Generate sample from $q_\theta$ via Langevin dynamics:}
    \FOR{sample step $k = 1$ to $K$}
    \STATE $\sx^k \gets \sx^{k-1} -  \nabla_\x E_\theta (\sx^{k-1}) + \omega, \;\; \omega \sim \mathcal{N}(0,\sigma)$
    \ENDFOR
    \STATE $\x^-_i = \Omega(\sx^k_i) $ \vspace{2mm} 
    \STATE \emph{$\triangleright$ Optimize objective $\alpha \mathcal{L}_{2} + \mathcal{L}_\text{ML}$ wrt $\theta$:} \vspace{1mm}
    \STATE $\Delta \theta \gets \nabla_\theta \frac{1}{N} \sum_i  \alpha ( E_\theta(\x^+_i)^2 + E_\theta(\x^-_i)^2) + E_\theta(\x^+_i) - E_\theta(\x^-_i) $
    \STATE Update $\theta$ based on $\Delta \theta$ using Adam optimizer  \vspace{2mm}
    \STATE $\mathcal{B} \gets \mathcal{B} \cup \sx_i$
    \ENDWHILE
  \end{algorithmic}
 \caption{Energy training algorithm}
 \end{algorithm}
    \end{minipage}\hfill
    \begin{minipage}{0.5\textwidth}
    \centering
    \includegraphics[width=0.9\linewidth]{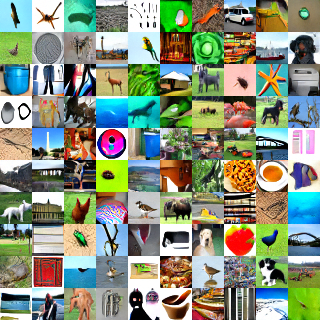}
    \caption{\small Conditional ImageNet32x32 EBM samples}
    \label{fig:cond_qual}
    \end{minipage}
    \vspace{-10pt}
\end{figure}

\section{Image Modeling}
\begin{figure}[h]
\begin{center}
\begin{subfigure}{0.23\textwidth}
    \centering
    \includegraphics[width=0.9\linewidth]{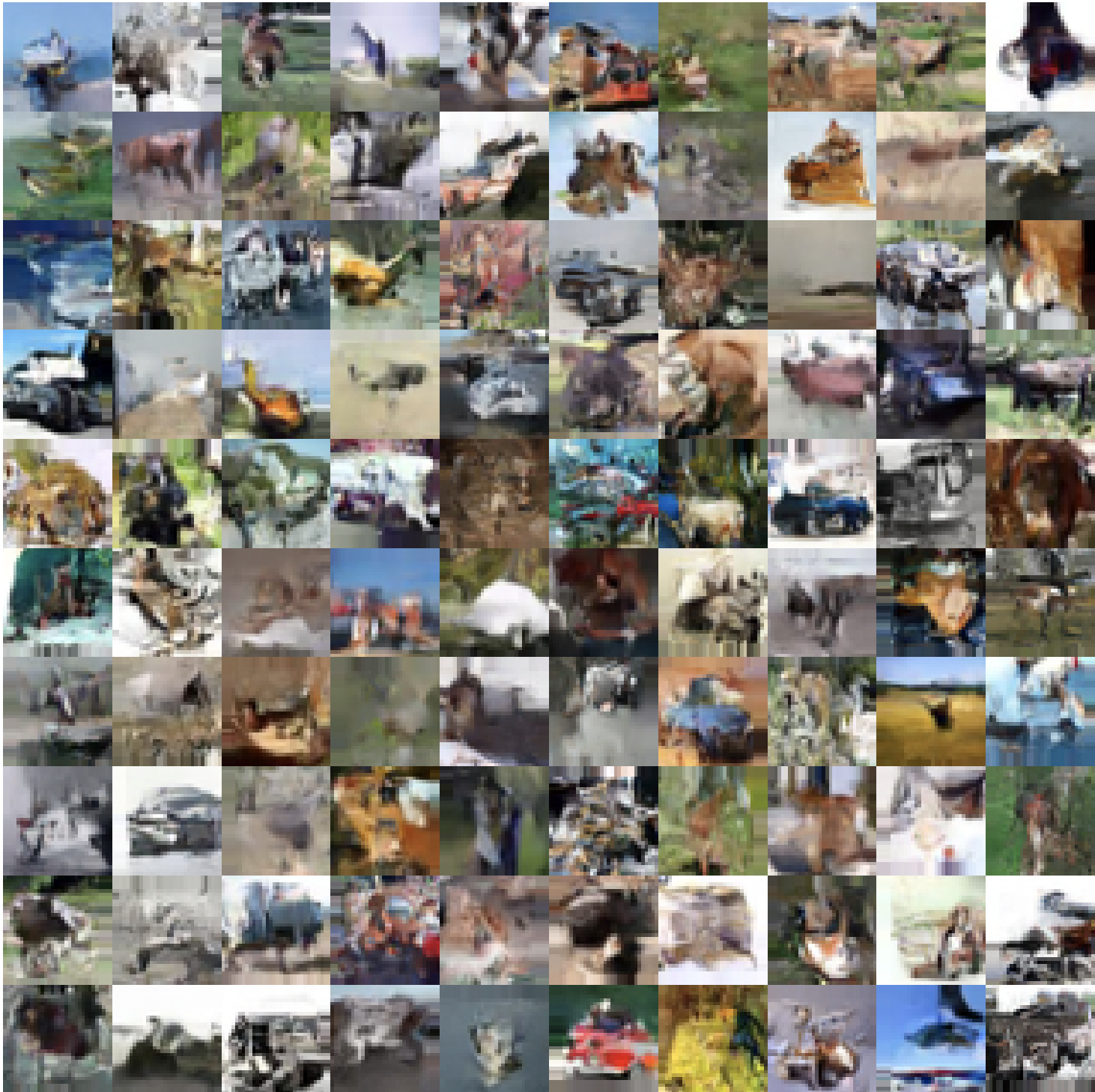}
    \caption{GLOW Model}
\end{subfigure}%
\begin{subfigure}{0.23\textwidth}
    \centering
    \includegraphics[width=0.9\linewidth]{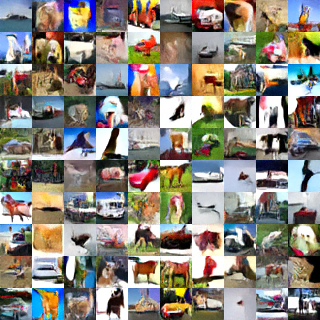}
    \caption{EBM}
\end{subfigure}%
\begin{subfigure}{0.23\textwidth}
    \centering
    \includegraphics[width=0.9\linewidth]{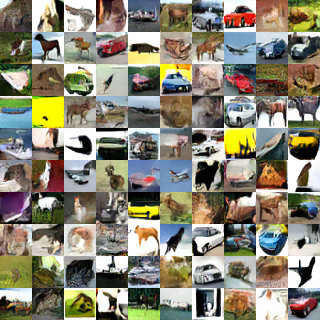}
    \centering
    \caption{EBM (10 historical)}
    \label{fig:hist}
\end{subfigure}
\begin{subfigure}{0.23\textwidth}
    \centering
    \includegraphics[width=0.9\linewidth]{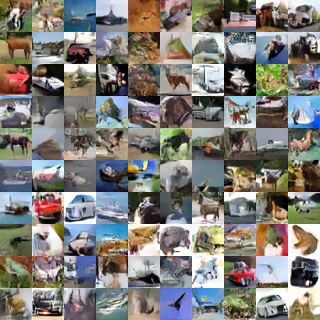}
    \centering
    \caption{EBM Sample Buffer}
    \label{fig:replay}
\end{subfigure}
\caption{\small Comparison of image generation techniques on unconditional CIFAR-10 dataset.}
\label{fig:qual_images}
\end{center}
\end{figure}

In this section, we show that EBMs are effective generative models for images. We show EBMs are able to generate high fidelity images and exhibit mode coverage on CIFAR-10 and ImageNet. We further show EBMs exhibit adversarial robustness and better out-of-distribution behavior than other likelihood models.  Our model is based on the ResNet architecture (using conditional gains and biases per class \citep{dumoulin2017learned} for conditional models) with details in the supplement. We present sensitivity analysis, likelihoods, and ablations in the supplement in A.4. We provide a comparison between EBMs and other likelihood models in A.5. Overall, we find that EBMs are both more parameter/computationally efficient than likelihood models, though worse than GANs.

\subsection{Image Generation}

We show unconditional CIFAR-10 images in \fig{fig:qual_images}, with comparisons to GLOW \citep{kingma2018glow}, and conditional ImageNet32x32 images in \fig{fig:cond_qual}.  We provide qualitative images of ImageNet128x128  and other visualizations in A.1.

\begin{figure}[H]
    \centering
    \begin{minipage}{0.6\textwidth}
\resizebox{\textwidth}{!}{
    \begin{tabular}{lcc}
    \toprule
    Model & Inception* & FID\\
    \midrule
    \textbf{CIFAR-10 Unconditional} & & \\
    \midrule
    PixelCNN \citep{VanOord2016Pixel} & 4.60 & 65.93  \\
    PixelIQN \citep{ostrovski2018autoregressive}& 5.29 & 49.46 \\
    EBM (single) & 6.02 & 40.58 \\
    DCGAN \citep{Radford2016Unsupervised} & 6.40  & 37.11 \\
    WGAN + GP \citep{Gulrajani2017Improved} &  6.50 &  36.4\\
    EBM (10 historical ensemble) & 6.78 & 38.2 \\
    SNGAN \citep{miyato2018spectral} & \textbf{8.22} & 21.7 \\
    \midrule
    \textbf{CIFAR-10 Conditional} & & \\
    \midrule    
    Improved GAN & 8.09 & - \\
    EBM (single) & 8.30 & 37.9 \\
    Spectral Normalization GAN & \textbf{8.59} &  25.5\\
    \midrule
    \textbf{ImageNet 32x32 Conditional} & & \\
    \midrule    
    PixelCNN & 8.33 & 33.27 \\
    PixelIQN & 10.18 & 22.99 \\
    EBM (single) & \textbf{18.22} & \textbf{14.31} \\
    \midrule
    \textbf{ImageNet 128x128 Conditional} & & \\
    \midrule    
    ACGAN \citep{odena2017conditional} & 28.5 & - \\
    EBM* (single) & 28.6 & 43.7 \\
    SNGAN & \textbf{36.8} & \textbf{27.62} \\
    
    \bottomrule

    \end{tabular}
}
    \caption{\small Table of Inception and FID scores for ImageNet32x32 and CIFAR-10. Quantitative numbers for ImageNet32x32 from \citep{ostrovski2018autoregressive}. (*) We use Inception Score (from original OpenAI repo) to compare with legacy models, but strongly encourage future work to compare soley with FID score, since Langevin Dynamics converges to minima that artificially inflate Inception Score. (**) conditional EBM models for 128x128 are smaller than those in SNGAN.}
    \label{tbl:inception}
    \end{minipage}\hfill
    \begin{minipage}{0.35\textwidth}
    \centering
    \includegraphics[width=1.0\linewidth]{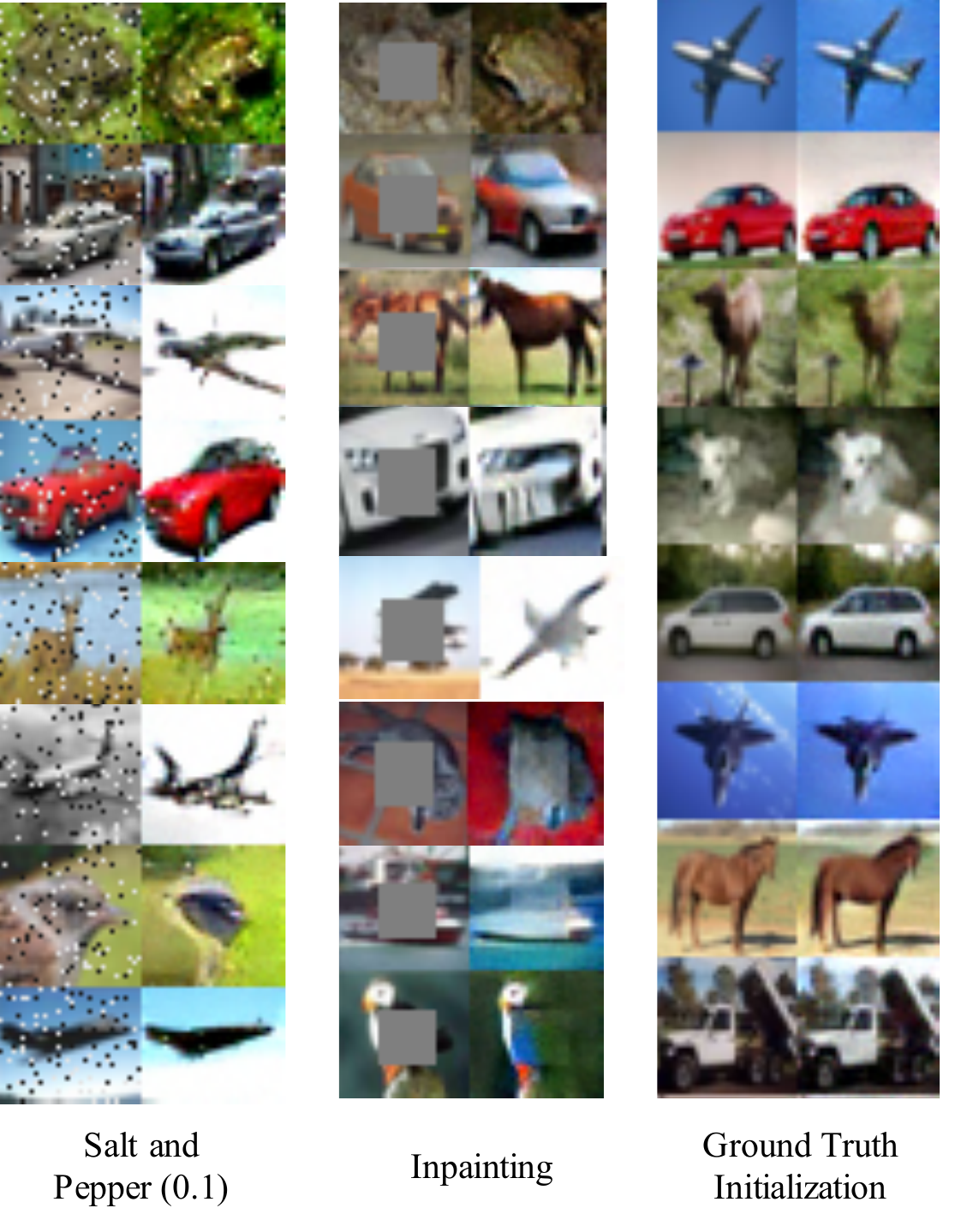}
    \caption{\small EBM image restoration on images in the \textbf{test} set via MCMC. The right column shows failure (approx. 10\% objects change with ground truth initialization and 30\% of objects change in salt/pepper corruption or in-painting. Bottom two rows shows worst case of change.)} 
    \label{fig:decorrupt}
    \end{minipage}
\end{figure}

We quantitatively evaluate image quality of EBMs with Inception score \citep{Salimans2016Improved} and FID score \citep{heusel2017gans} in \tbl{tbl:inception}. Overall we obtain significantly better scores than likelihood models PixelCNN and PixelIQN, but worse than SNGAN \citep{miyato2018spectral}. We found that in the unconditional case, mode exploration with Langevin took a very long time, so we also experimented in 
\emph{EBM (10 historical ensemble)} with sampling joint from the last 10 snapshots of the model. At training time, extensive exploration is ensured with the replay buffer (\fig{fig:replay}). Our models have similar number of parameters to SNGAN, but we believe that significantly more parameters may be necessary to generate high fidelity images with mode coverage. On ImageNet128x128, due to computational constraints, we train a smaller network than SNGAN and do not train to convergence.


\subsection{Mode Evaluation}

We evaluate over-fitting and mode coverage in EBMs. To test over-fitting, we plotted histogram of energies for CIFAR-10 train and test dataset in \fig{fig:svhn_hist} and note almost identical curves. In the supplement, we show that the nearest neighbor of generated images are not identical to images in the training dataset. To test mode coverage in EBMs, we investigate MCMC sampling on corrupted CIFAR-10 test images. Since Langevin dynamics is known to mix slowly \citep{Neal2011MCMC} and reach local minima, we believe that good denoising after limited number of steps of sampling indicates probability modes at respective test images. Similarly, lack of movement from a ground truth test image initialization after the same number of steps likely indicates probability mode at the test image.  In \fig{fig:decorrupt}, we find that if we initialize sampling with images from the test set, images do not move significantly. However, under the same number of steps,  \fig{fig:decorrupt} shows that we are able to reliably decorrupt masked and salt and pepper corrupted images, indicating good mode coverage. We note that large number of steps of sampling lead to more saturated images, which are due to sampling low temperature modes, which are saturated across likelihood models (see appendix). In comparison, GANs have been shown to miss many modes of data and cannot reliably reconstruct many different test images \citep{yeh2017semantic}. We note that such decorruption behavior is a nice property of implicit generation without need of explicit knowledge of corrupted pixels. 


\begin{figure}[H]
    \centering
    \begin{minipage}{0.45\textwidth}
    \centering
    \includegraphics[width=1.0\linewidth]{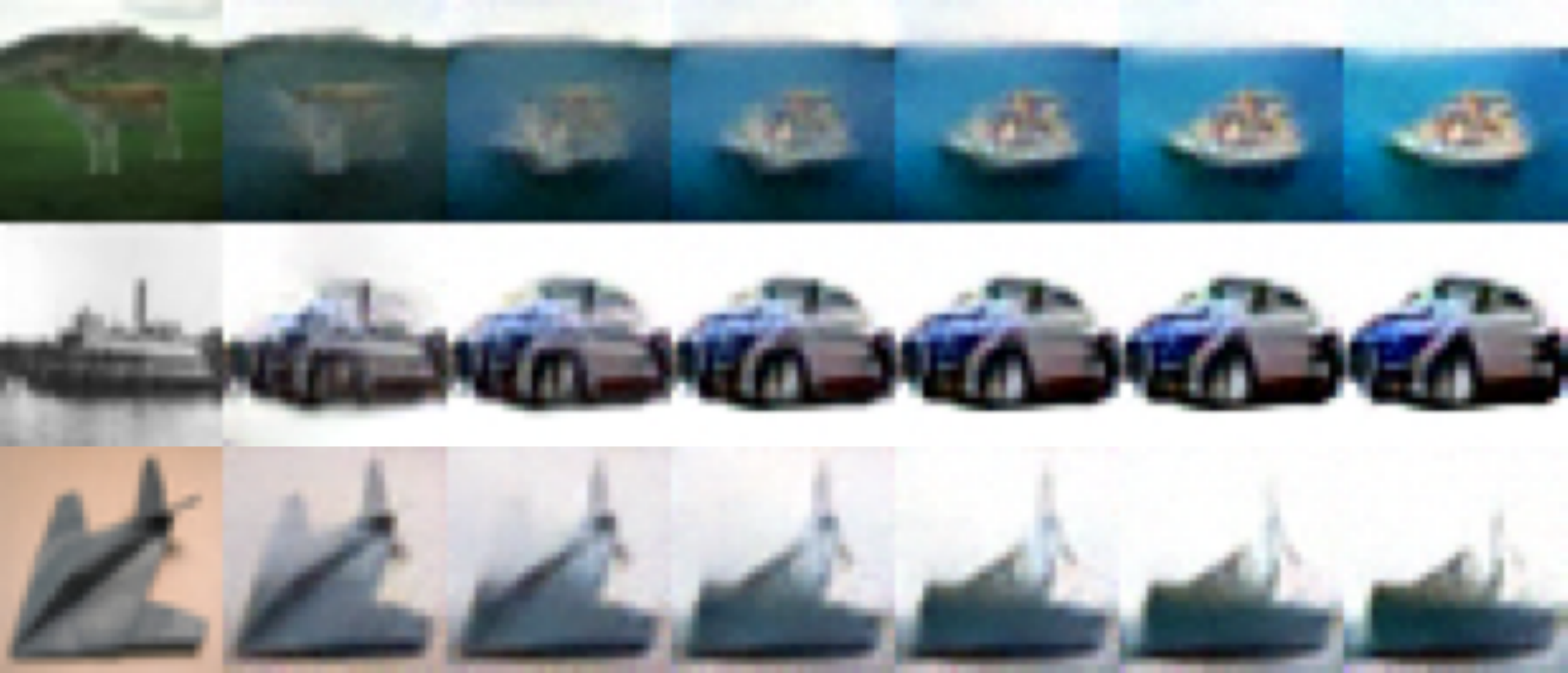}
    \caption{\small Illustration of cross-class implicit sampling on a conditional EBM. The EBM is conditioned on a particular class but is initialized with an image from a separate class.} 
    \label{fig:cross_class}
    \end{minipage}\hfill
    \begin{minipage}{0.5\textwidth}
    \centering
    \includegraphics[width=1.0\linewidth]{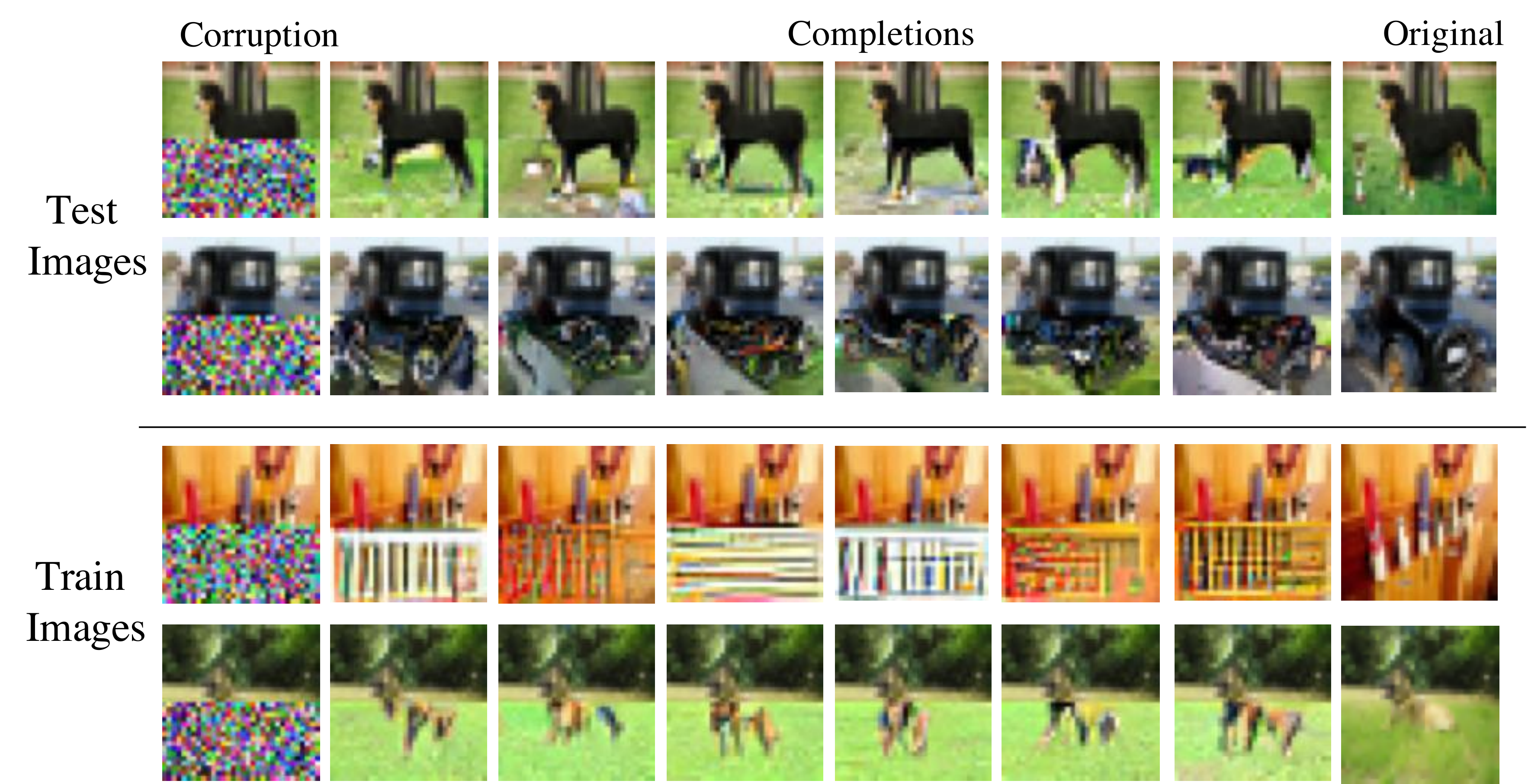}
    \caption{\small Illustration of image completions on conditional ImageNet model. Our models exhibit diversity in inpainting.} 
    \label{fig:completions}
    \end{minipage}
    \vspace{-10pt}
\end{figure}

Another common test for mode coverage and overfitting is masked inpainting \citep{VanOord2016Pixel}. In \fig{fig:completions}, we mask out the bottom half of ImageNet images and test the ability to sample the masked pixels, while fixing the value of unmasked pixels. Running Langevin dynamics on the images, we find diversity of completions on train/test images, indicating low overfitting on training set and diversity characterized by likelihood models. Furthermore initializing sampling of a class conditional EBM with images from images from another class, we can further test for presence of probability modes at images far away from the those seen in training.  We find in \fig{fig:cross_class} that sampling on such images using an EBM is able to generate images of the target class, indicating semantically meaningful modes of probability even far away from the training distribution.



\subsection{Adversarial Robustness}

\begin{wrapfigure}{l}{0.5\textwidth}
\begin{center}
\begin{subfigure}{0.22\textwidth}
\includegraphics[width=1.0\linewidth]{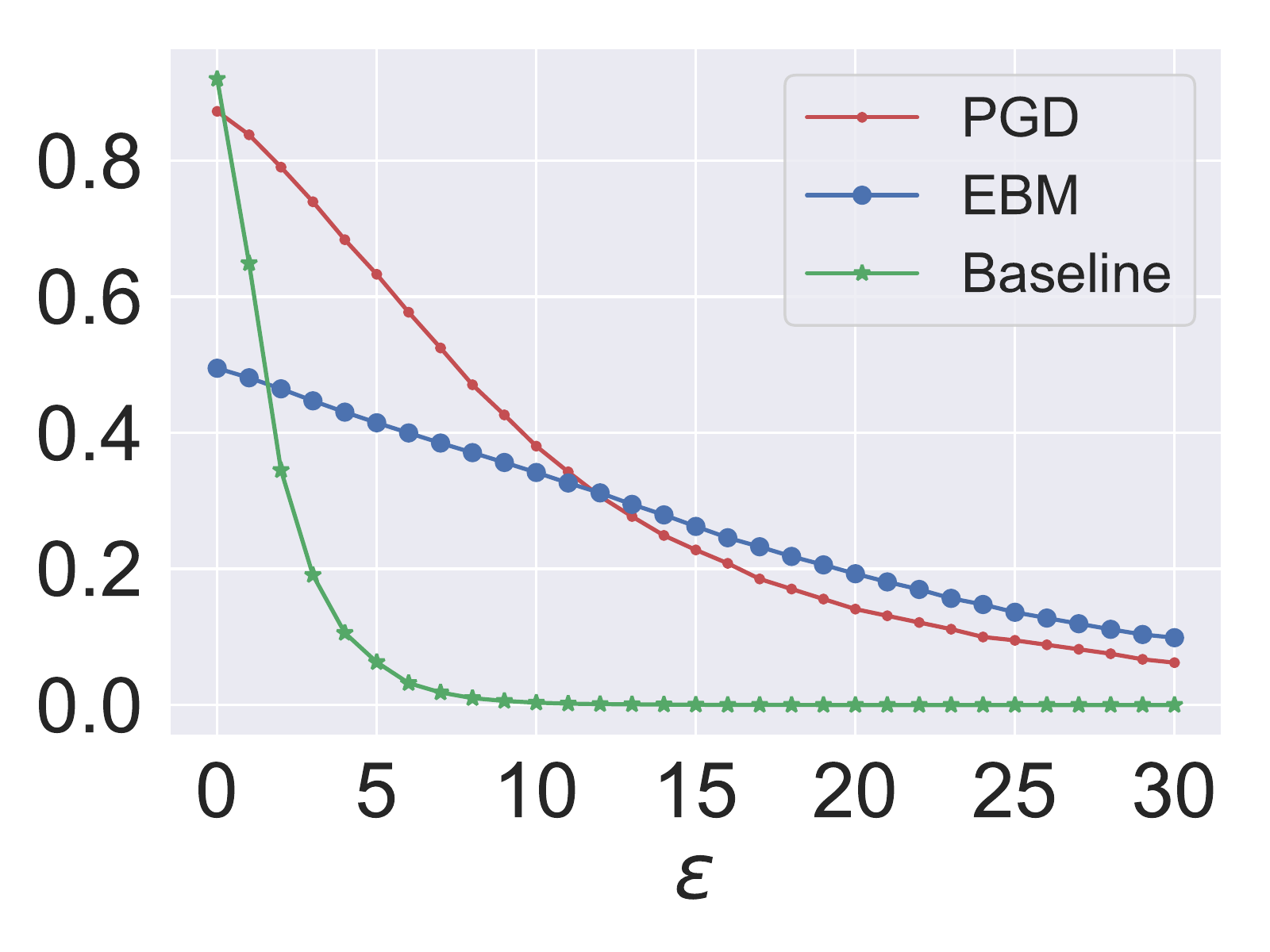}
\caption{$L_{\infty}$ robustness}
\end{subfigure} 
\begin{subfigure}{0.22\textwidth}
\includegraphics[width=1.0\linewidth]{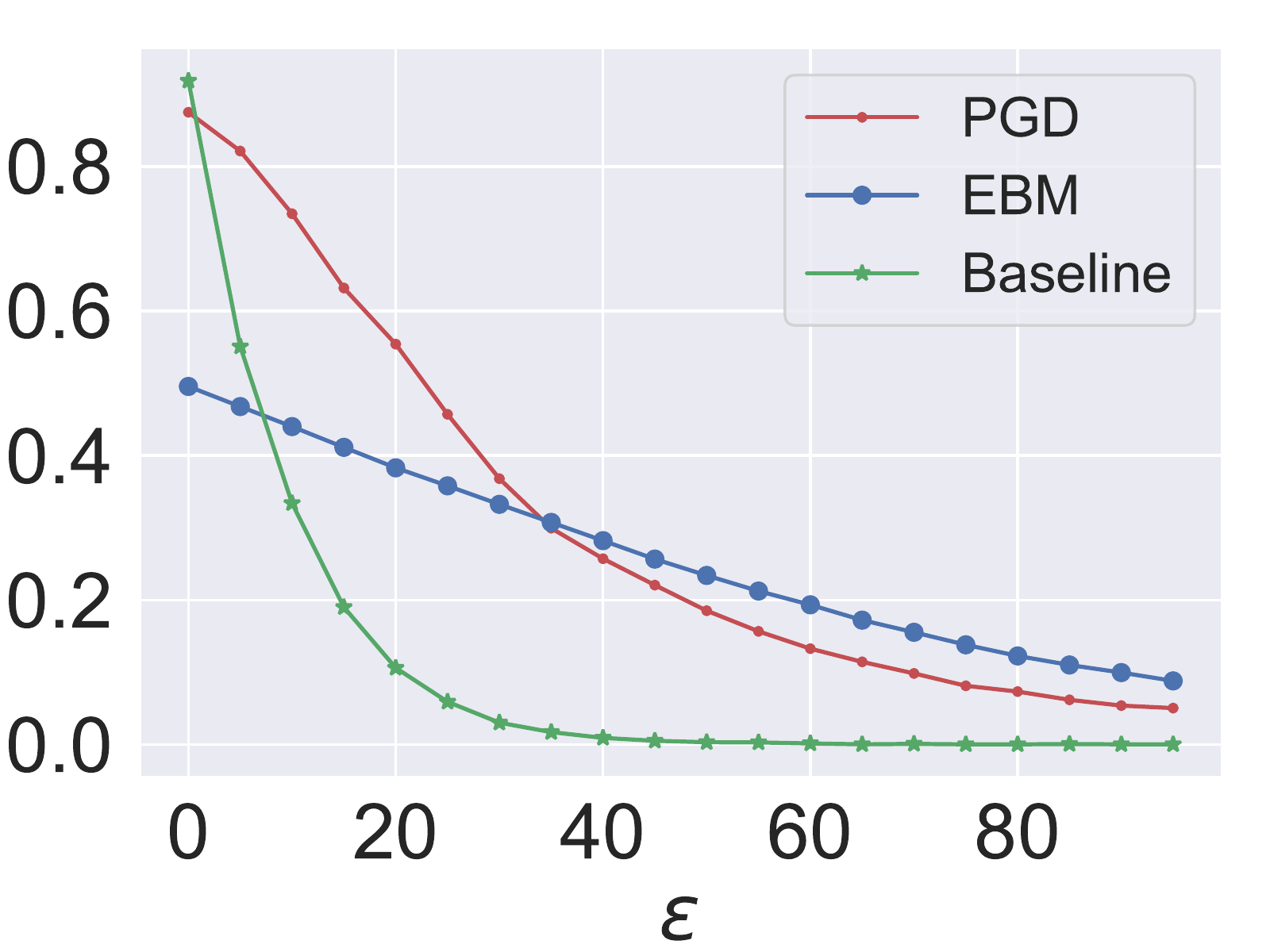}

\caption{$L_2$ Robustness}
\end{subfigure}
\end{center}
\caption{\small $\epsilon$ plots under $L_{\infty}$ and $L_2$ attacks of conditional EBMs as compared to PGD trained models in \citep{madry2017towards} and  a baseline Wide ResNet18.}
\label{fig:adversarial}
\end{wrapfigure}

We show conditional EBMs exhibit adversarial robustness on CIFAR-10 classification, \textbf{without explicit} adversarial training. To compute logits for classification, we compute the negative energy of the image in each class. Our model, without fine-tuning, achieves an accuracy of 49.6\%. \fig{fig:adversarial} shows adversarial robustness curves. We ran 20 steps of PGD as in \citep{madry2017towards}, on the above logits. To undergo classification, we then ran 10 steps sampling initialized from the starting image (with a bounded deviation of 0.03) from each conditional model, and then classified using the lowest energy conditional class. We found that running PGD incorporating sampling was less successful than without. Overall we find in  \fig{fig:adversarial} that EBMs are very robust to adversarial perturbations and outperforms the SOTA $L_{\infty}$ model  in \citep{madry2017towards} on $L_{\infty}$ attacks with $\epsilon > 13$.
\subsection{Out-of-Distribution Generalization}

\begin{figure}[H]
    \centering
    \begin{minipage}{0.3\textwidth}
    \centering
    \includegraphics[width=1.0\linewidth]{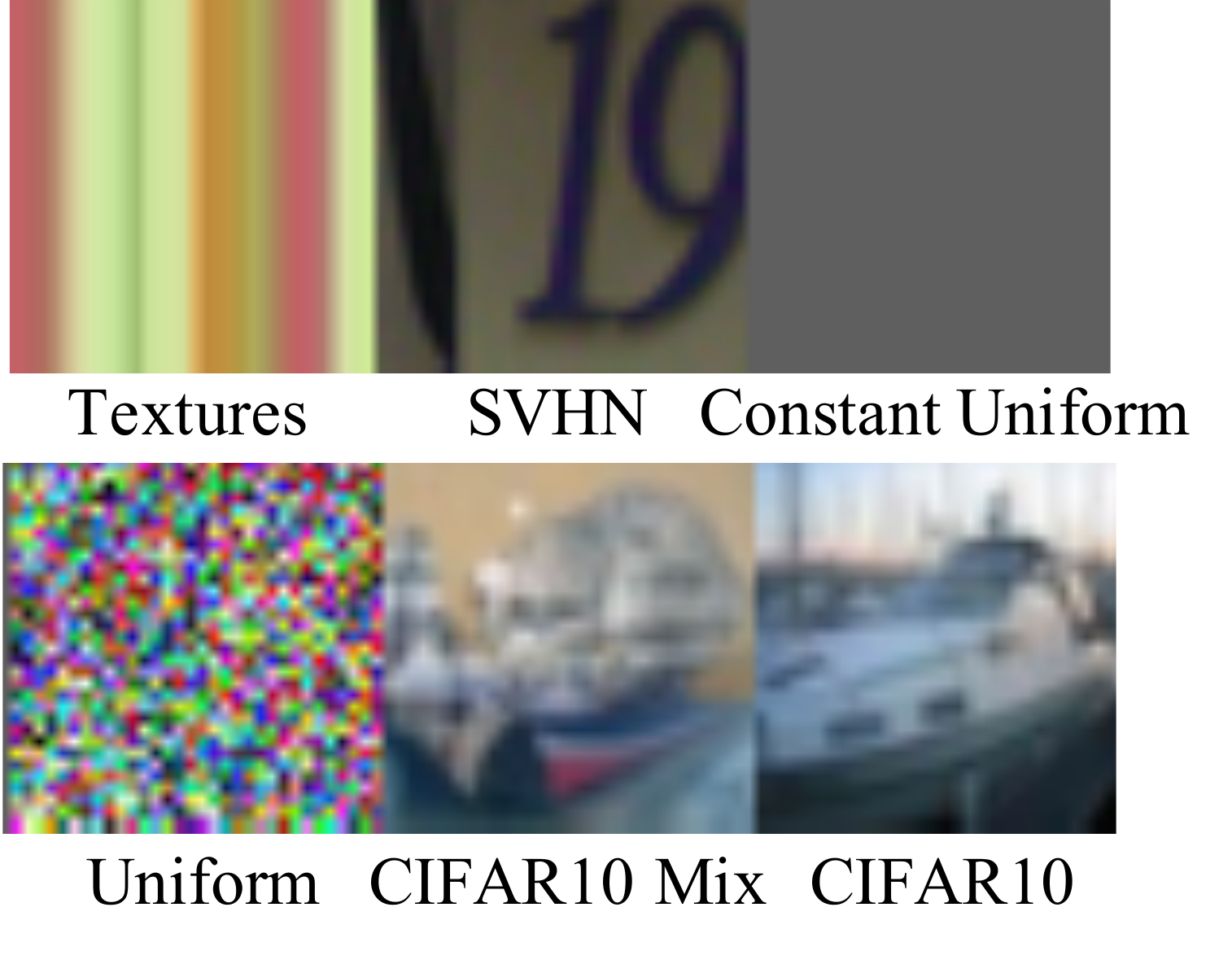}
    
    \caption{Illustration of images from each of the out of distribution dataset.} 
    \label{fig:decker}
    \end{minipage}\hfill
    \begin{minipage}{0.65\textwidth}
    \vspace{-10pt}
    \begin{center}
    \begin{tabular}{lccc}
        \toprule
        Model & PixelCNN++ & Glow & EBM (ours) \\
        \midrule
        SVHN & 0.32 & 0.24 & \textbf{0.63}\\
        Textures & 0.33 & 0.27 & \textbf{0.48}\\
        Constant Uniform & 0.0 & 0.0 & \textbf{0.30}\\
        Uniform & 1.0 & 1.0 & \textbf{1.0} \\
        CIFAR10 Interpolation & \textbf{0.71} & 0.59 & 0.70 \\
        Average & 0.47 & 0.42 & \textbf{0.62} \\
        \bottomrule
    \end{tabular}
    \caption{\small AUROC scores of out of distribution classification on different datasets. Only our model gets better than chance classification.}
    \label{tbl:generalization}
    \end{center}
    \end{minipage}
\end{figure}

We show EBMs exhibit better out-of-distribution (OOD) detection than other likelihood models. Such a task requires models to have high likelihood on the data manifold and low likelihood at all other locations and can be viewed as a proxy of log likelihood. Surprisingly, \citet{anonymous2019do} found likelihood models such as VAE, PixelCNN, and Glow models, are unable to distinguish data assign higher likelihood to many OOD images. We constructed our OOD metric following following \citep{hendrycks2016baseline} using Area Under the ROC Curve (AUROC) scores computed based on classifying CIFAR-10 test images from other OOD images using relative log likelihoods. We use SVHN, Textures \citep{cimpoi14describing}, monochrome images, uniform noise and interpolations of separate CIFAR-10 images as OOD distributions. We provide examples of OOD images in \fig{fig:decker}. We found that our proposed OOD metric correlated well with training progress in EBMs.

\begin{figure*}
\begin{center}
\begin{subfigure}{0.25\textwidth}
\includegraphics[width=1.0\linewidth]{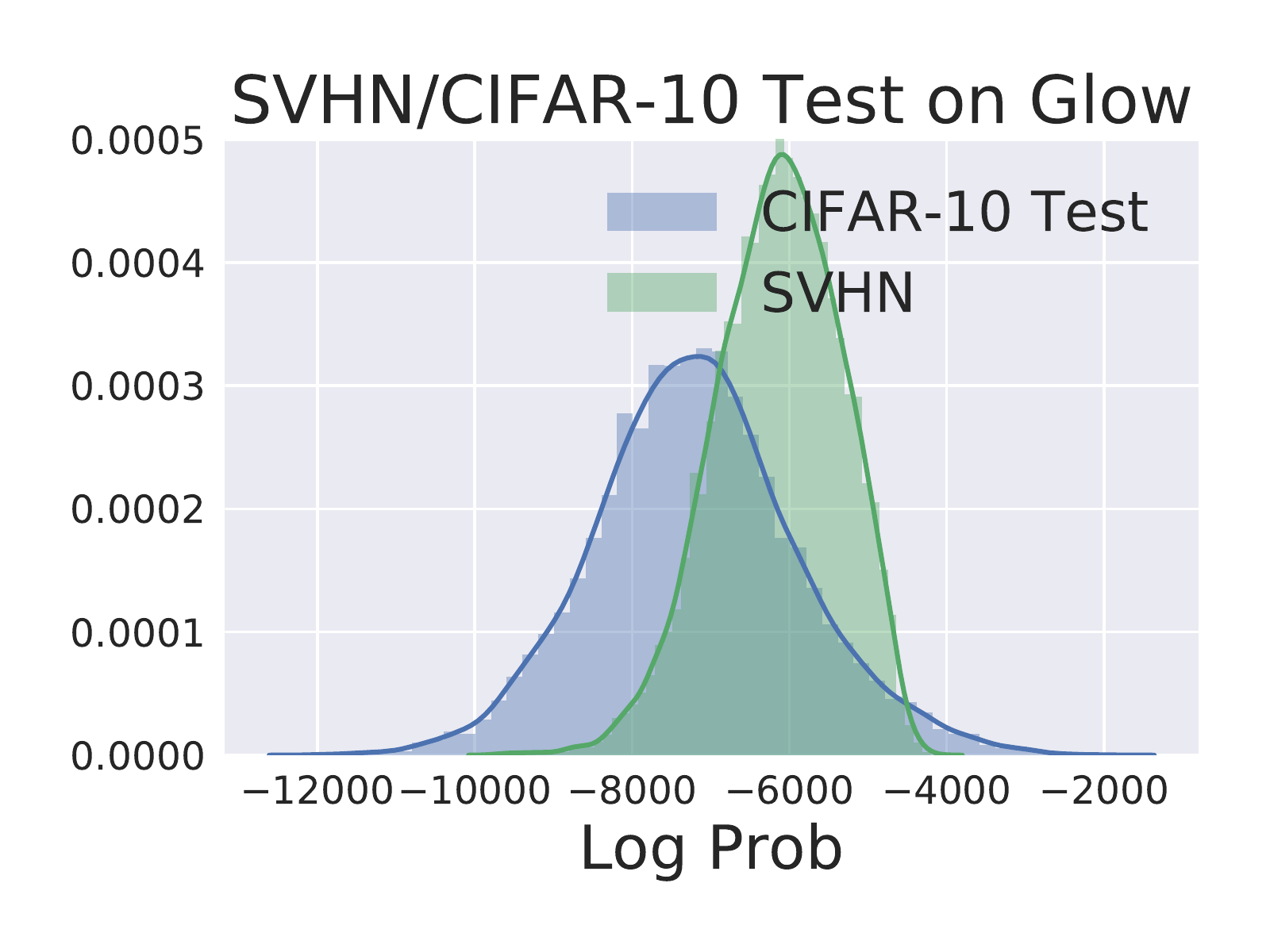}
\end{subfigure}%
\begin{subfigure}{0.25\textwidth}
\includegraphics[width=1.0\linewidth]{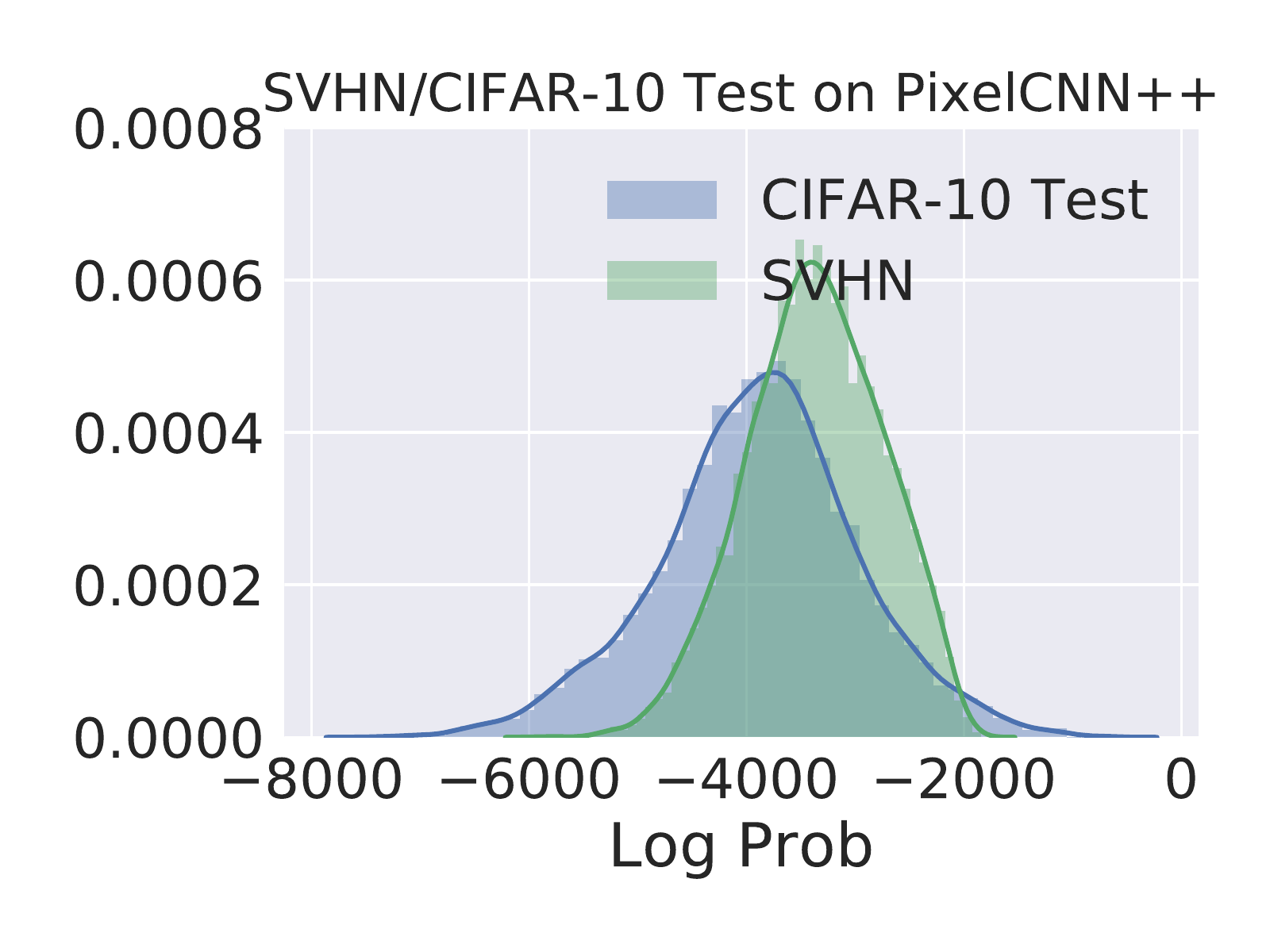}
\end{subfigure}%
\begin{subfigure}{0.25\textwidth}
\includegraphics[width=1.0\linewidth]{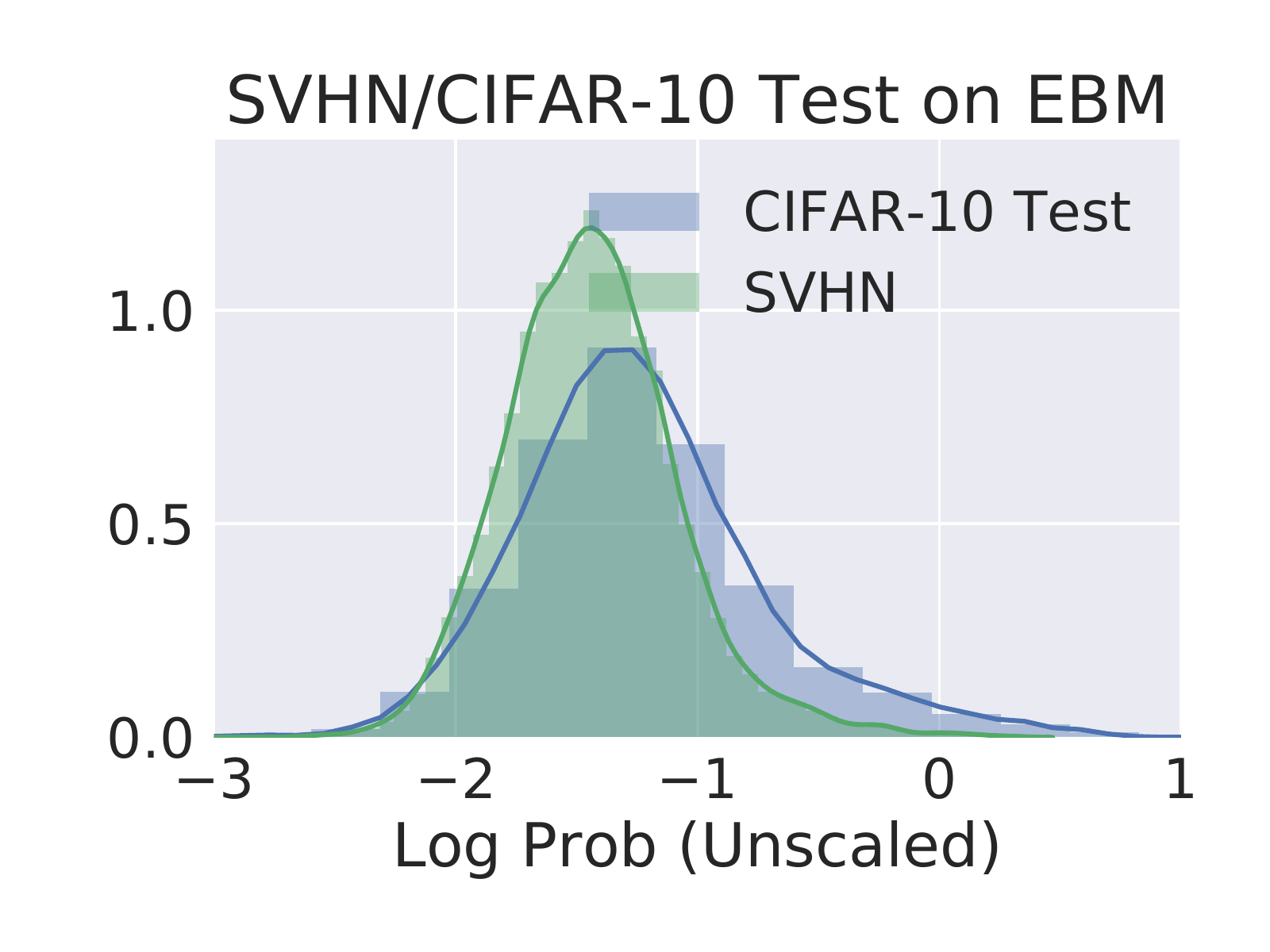}
\end{subfigure}%
\begin{subfigure}{0.25\textwidth}
\includegraphics[width=1.0\linewidth]{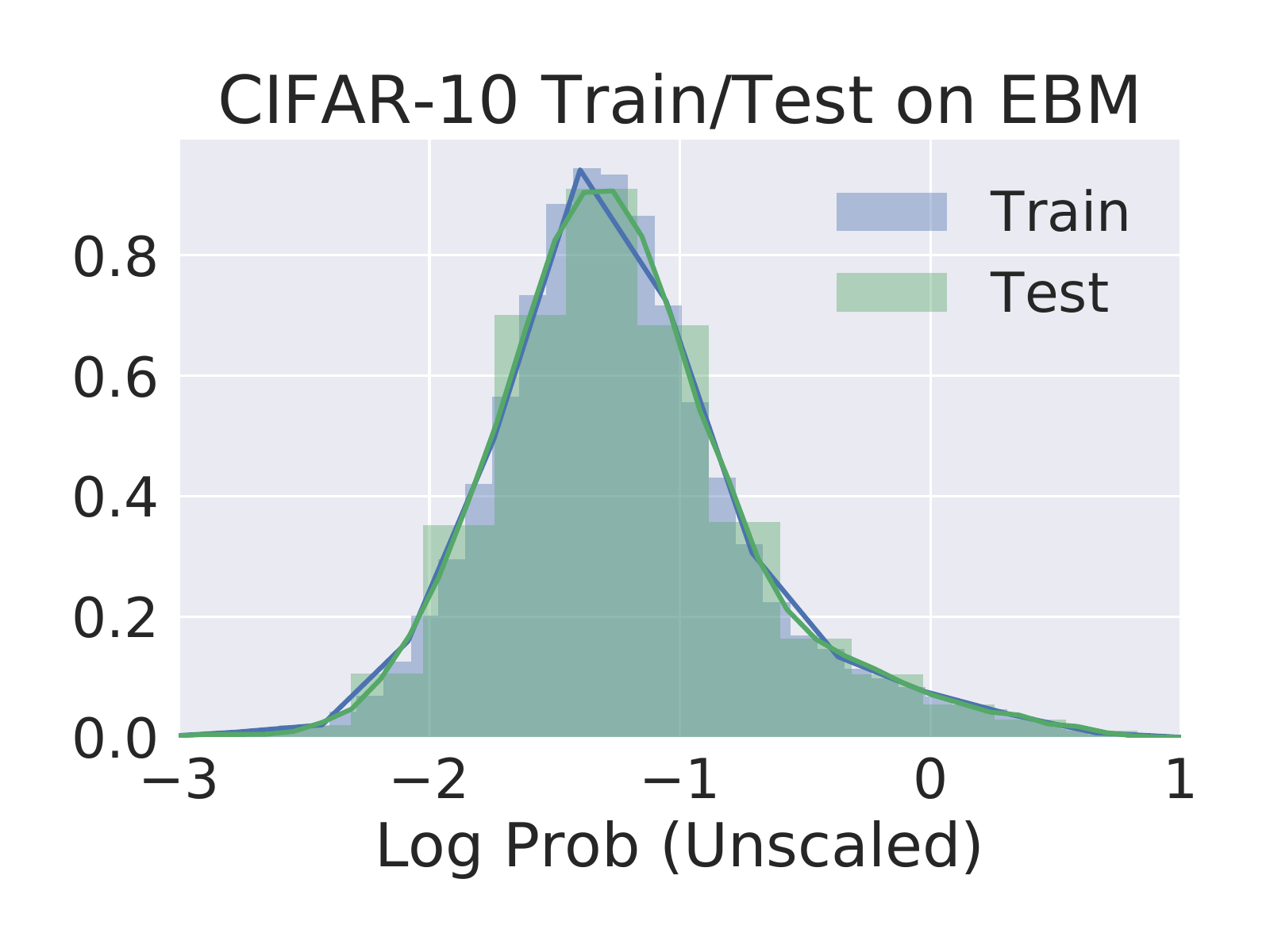}
\end{subfigure}
\end{center}
\caption{\small Histogram of relative likelihoods for various datasets for Glow, PixelCNN++ and EBM models}
\label{fig:svhn_hist}
\end{figure*}

In \tbl{tbl:generalization}, unconditional EBMs perform \textbf{significantly better} out-of-distribution than other auto-regressive and flow generative models and have OOD scores of 0.62 while the closest, PixelCNN++, has a OOD score of 0.47. We provide histograms of relative likelihoods for SVHN in \fig{fig:svhn_hist} which are also discussed in \citep{anonymous2019do, hendrycks2018oe}. We believe that the reason for better generalization is two-fold. First, we believe that the negative sampling procedure in EBMs helps eliminate spurious minima. Second, we believe EBMs have a flexible structure that allows global context when estimating probability without imposing constraints on latent variable structure. In contrast, auto-regressive models model likelihood sequentially, which makes global coherence difficult. In a different vein, flow based models must apply continuous transformations onto a continuous connected probability distribution which makes it very difficult to model disconnected modes, and thus assign spurious density to connections between modes.



\section{Trajectory Modeling}

We show that EBMs generate and generalize well in the different domain of trajectory modeling. We train EBMs to model dynamics of a simulated robot hand manipulating a free cube object \citep{openaicube}. We generated 200,000 different trajectories of length 100, from a trained policy (with every 4th action set to a random action for diversity), with a 90-10 train-test split. Models are trained to predict positions of all joints in the hand and orientation and position of the cube one step in the future. We test performance by evaluating many step roll-outs of self-predicted trajectories. 

\subsection{Training Setup and Metrics}


We compare EBM models to feedforward models (FC), both of which are composed of 3 layers of 128 hidden units. We apply spectral normalization to FC to prevent multi-step explosion. We evaluate multi-step trajectories by computing Frechet Distance \citep{dowson1982frechet} between predicted and ground distributions across all states at timestep $t$. We found this metric was a better metric of trajectories than multi-step MSE due to accumulation of error.

\begin{figure}[H]
    \centering
    \begin{minipage}{0.45\textwidth}
        \includegraphics[width=1.0\linewidth]{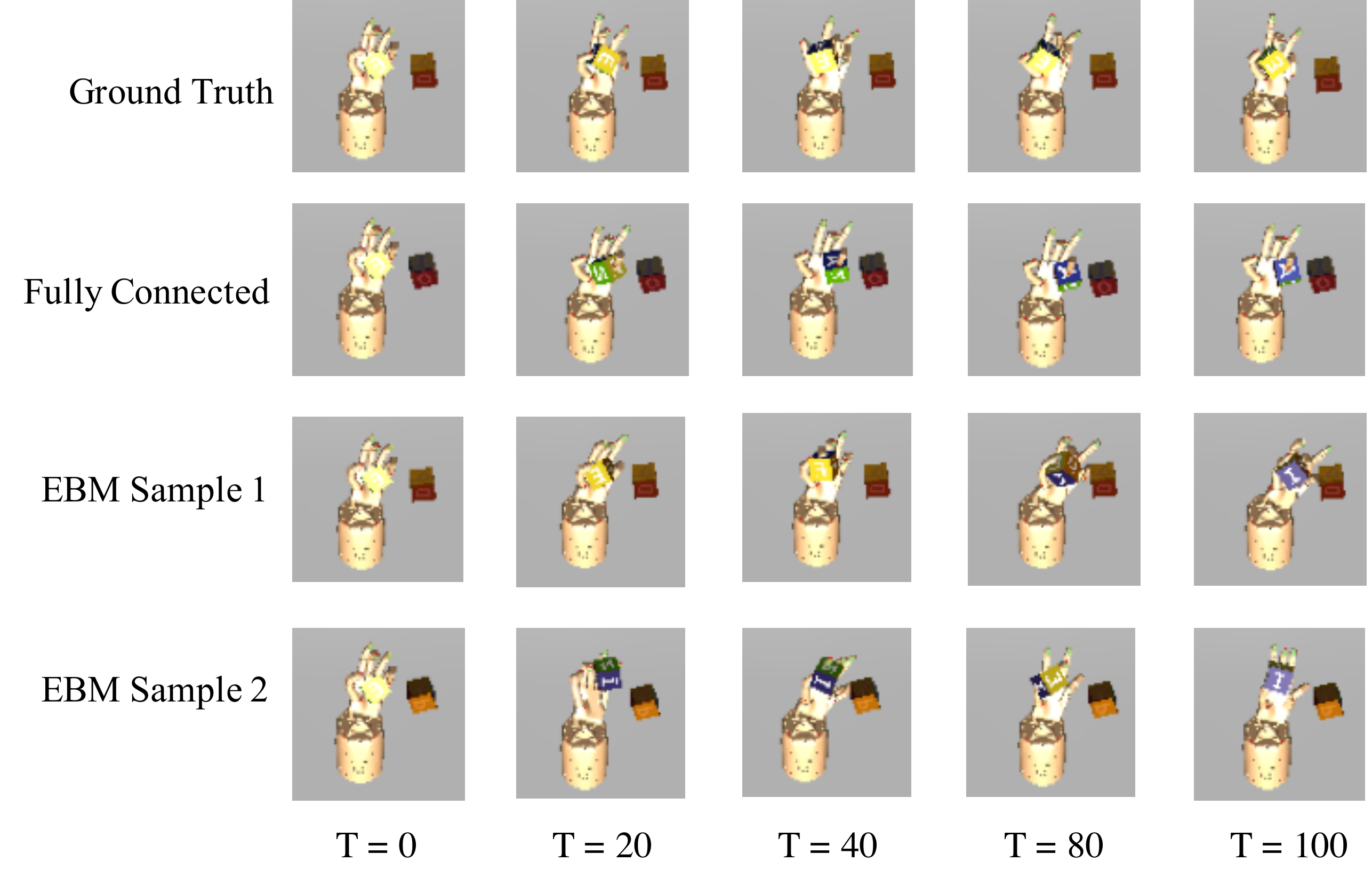}
        \caption{\small Views of hand manipulation trajectories generated unconditionally from the same state(1st frame).}
        \label{fig:traj_images}
    \end{minipage}\hfill
    \begin{minipage}{0.45\textwidth}
        \centering
        \includegraphics[width=1.0\linewidth]{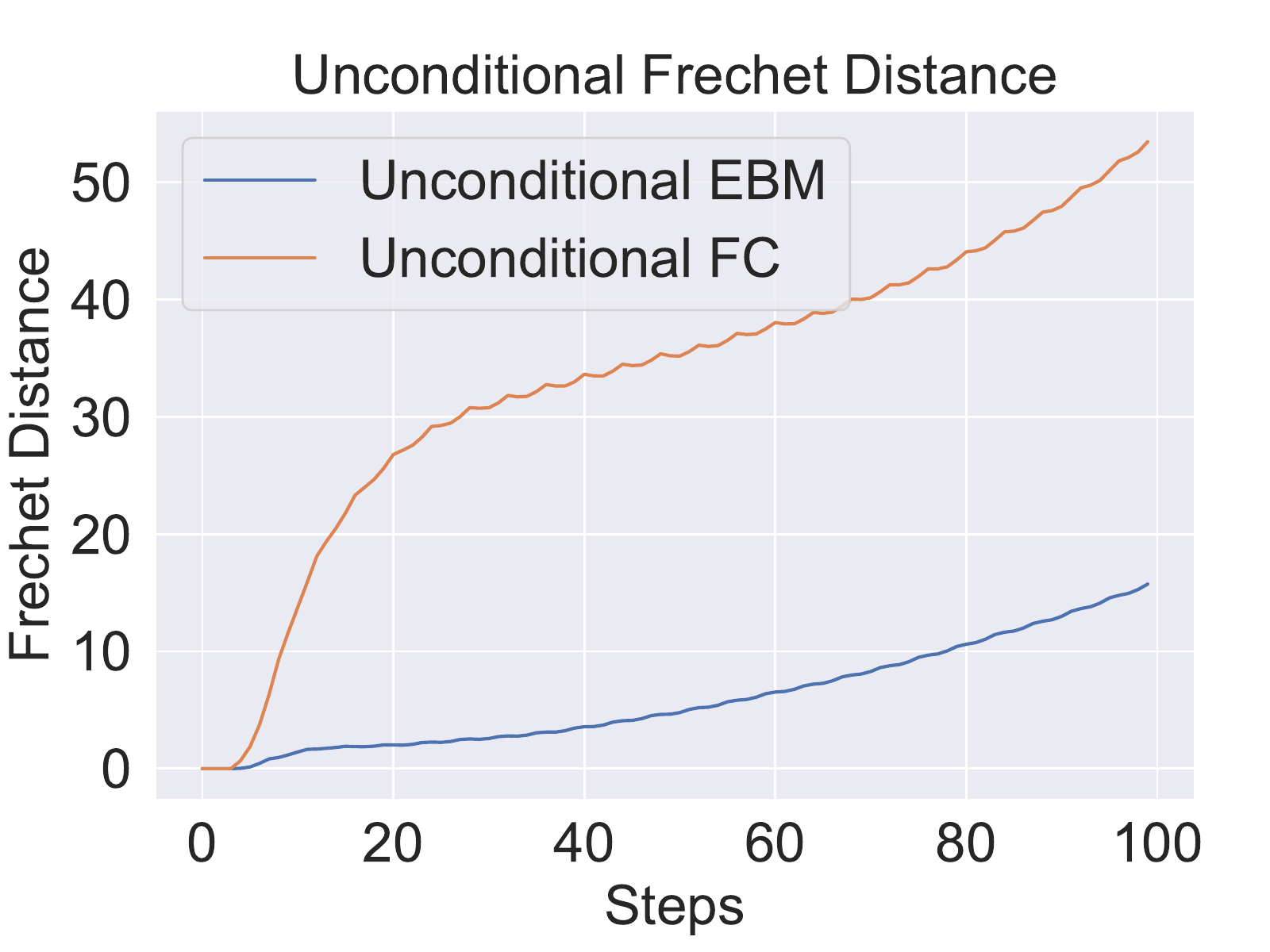}
        \caption{\small Conditional and Unconditional Modeling of Hand Manipulation through Frechet Distance}
        \label{fig:cond_uncond_hand}
    \end{minipage}
    \vspace{-10pt}
\end{figure}



\subsection{Multi-Step Trajectory Generation}

We evaluated EBMs for both action conditional and unconditional prediction of multi-step rollouts. Quantitatively, by computing the average Frechet distance across all time-steps, unconditional EBM have value 5.96 while unconditional FC networks have a value of 33.28. Conditional EBM have value 8.97 while a conditional FC has value 19.75.  We provide plots of Frechet distance over time in \fig{fig:cond_uncond_hand}. In \fig{fig:cond_uncond_hand}, we observe that for unconditional hand modeling in a FC network, the Frechet distance increases dramatically in the first several time steps. Qualitatively, we found that the same FC networks stop predicting hand movement after several several steps as demonstrated in \fig{fig:traj_images}. In contrast, Frechet distance increases slowly for unconditional EBMs. The unconditional models are able to represent multimodal transitions such as different types of cube rotation and \fig{fig:traj_images} shows that the unconditional EBMs generate diverse realistic trajectories.

\section{Online Learning}

\begin{wraptable}{l}{0.5\textwidth}
\centering
\begin{tabular}{lc}
    \toprule
    Method & Accuracy \\
    \midrule
    EWC \citep{kirkpatrick2017overcoming} & 19.80 (0.05)\\
    SI \citep{zenke2017continual} & 19.67 (0.09)\\
    NAS \citep{schwarz2018progress} & 19.52 (0.29)\\
    LwF \citep{li2018learning}& 24.17 (0.33)\\
    VAE & 40.04 (1.31)\\
    EBM (ours) & \textbf{64.99} (4.27)\\
    \bottomrule
\end{tabular}
\caption{\small Comparison of various continual learning benchmarks. Values averaged acrossed 10 seeds reported as mean (standard deviation).}
\end{wraptable}
We find that EBMs also perform well in continual learning. We evaluate incremental class learning on the Split MNIST task proposed in \citep{farquhar2018towards}. The task evaluates overall MNIST digit classification accuracy given 5 sequential training tasks of disjoint pairs of digits.  We train a conditional EBM with 2 layers of 400 hidden units work and compare with a generative conditional VAE baseline with both encoder/decoder having 2 layers of 400 hidden units. Additional training details are covered in the appendix. We train the generative models to represent the joint distribution of images and labels and classify based off the lowest energy label. \citet{hsu2018re} analyzed common continual learning algorithms such as EWC \citep{kirkpatrick2017overcoming}, SI \citep{zenke2017continual} and NAS \citep{schwarz2018progress} and find they obtain performance around 20\%. LwF \citep{li2018learning} performed the best with performance of $24.17 \pm 0.33$ , where all architectures use 2 layers of 400 hidden units. However, since each new task introduces two new MNIST digits, a test accuracy of around 20\% indicates complete forgetting of previous tasks. In contrast, we found continual EBM training obtains \textbf{significantly higher} performance of $64.99 \pm 4.27$. All experiments were run with 10 seeds.

A crucial difference is that negative training in EBMs only locally "forgets" information corresponding to negative samples. Thus, when new classes are seen, negative samples are conditioned on the new class, and the EBM only forgets unlikely data from the new class. In contrast, the cross entropy objective used to train common continual learning algorithms down-weights the likelihood of all classes not seen.  We can apply this insight on other generative models, by maximizing the likelihood of a class conditional model at train time and then using the highest likelihood class as classification results. We ran such a baseline using a VAE and obtained a performance of $40.04 \pm 1.31$, which is higher than other continual learning algorithms but less than that in a EBM.

\section{Compositional Generation}
\begin{wrapfigure}{l}{0.5\textwidth}
\vspace{-15pt}
\begin{center}
\includegraphics[width=0.8\linewidth]{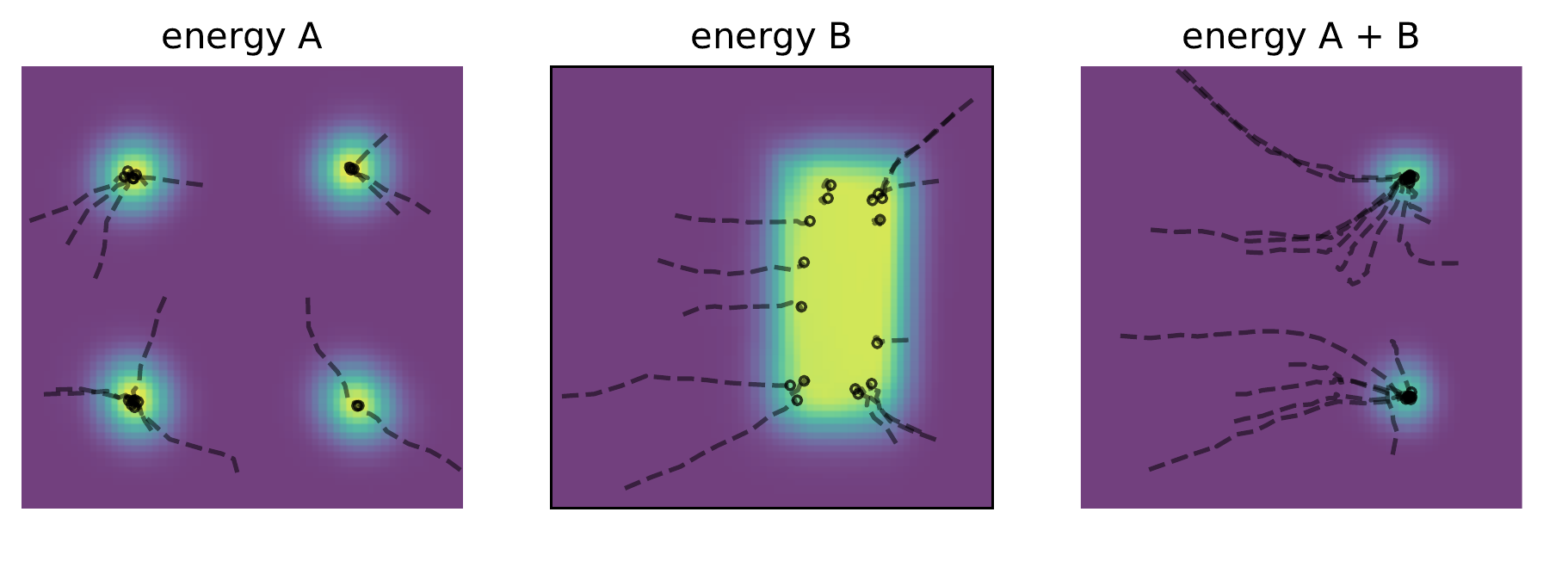}
\caption{\small A 2D example of combining EBMs through summation and the resulting sampling trajectories.} 
\label{fig:toy_comp}
\end{center}
\vspace{-10pt}
\end{wrapfigure}

Finally, we show compositionality through implicit generation in EBMs. Consider a set of conditional EBMs for separate independent latents.  Sampling through the joint distribution on all latents is represented by generation on an EBM that is the sum of each conditional EBM \citep{hinton1999products} and corresponds to a product of experts model. As seen in \fig{fig:toy_comp}, summation naturally allows composition of EBMs. We sample from joint conditional distribution through Langevin dynamics sequentially from each model.

We conduct our experiments on the dSprites dataset \citep{Higgins2017Beta}, which consists of all possible images of an object (square, circle or heart) varied by scale, position, rotation with labeled latents.  We trained conditional EBMs for each latent and found that scale, position and rotation worked well. The latent for shape was learned poorly, and we found that even our unconditional models were not able to reliably generate different shapes which was also found in \citep{Higgins2017Beta}. We show some results on CelebA in A.6.

\begin{figure}[H]
    \centering
    \begin{minipage}{0.45\textwidth}
        \begin{center}
        \includegraphics[width=0.8\linewidth]{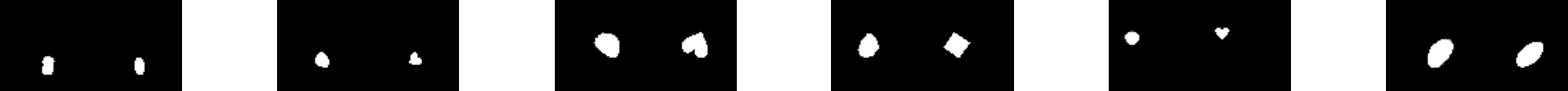}
        \caption{\small Samples from joint distribution of 4 independent conditional EBMs on scale, position, rotation and shape (left panel) with associated ground truth rendering (right panel).}
        \label{fig:joint}
        \end{center}
    \end{minipage}\hfill
    \begin{minipage}{0.45\textwidth}
    \centering
    \includegraphics[width=0.9\linewidth]{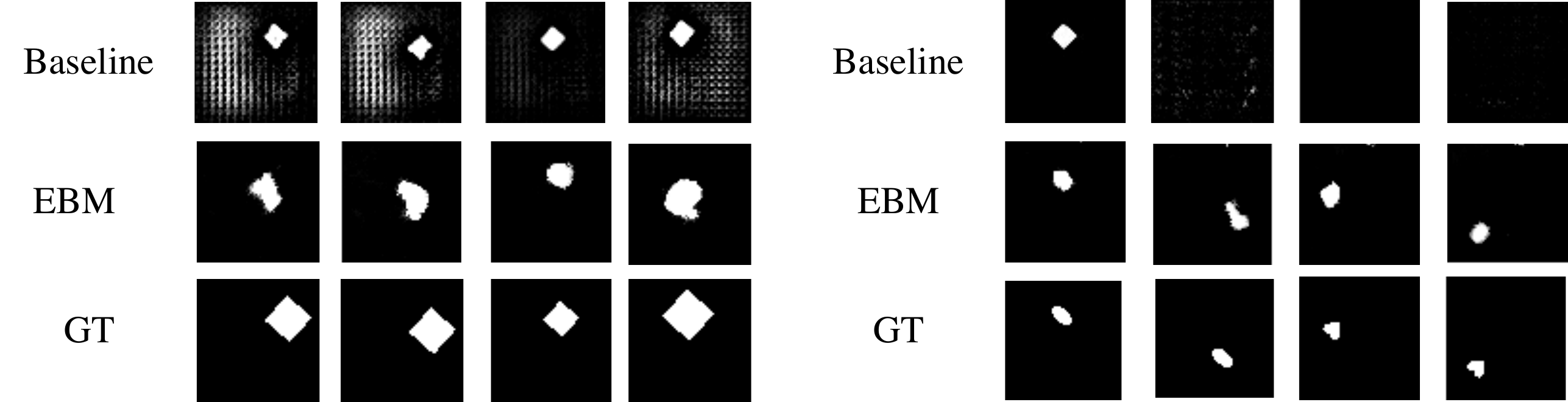}
    \caption{\small GT = Ground Truth. Images of cross product generalization of size-position (left panel) and shape-position (right panel).}
    \label{fig:comb_im}
    \end{minipage}
    \vspace{-10pt}
\end{figure}


\myparagraph{Joint Conditioning} In \fig{fig:joint}, we provide generated images from joint conditional sampling. Under such sampling we are able to generate images very close to ground truth for all classes with exception of shape. This result also demonstrates mode coverage across all data.

\myparagraph{Zero-Shot Cross Product Generalization}
We evaluate the ability of EBMs to generalize to novel combinations of latents. We generate three datasets, D1: different size squares at a central position, D2: smallest size square at each location, D3: different shapes at the center position. We evaluate size-position generalization by training independent energy functions on D1 and D2, and test on generating different size squares at all positions. We similarly evaluate shape-position generalization for D2 and D3. We generate samples at novel combinations by sampling from the summation of energy functions (we first finetune the summation energy to generate both training datasets using a KL term defined in the appendix). We compare against a joint conditional model baseline.

We present results of generalization in \fig{fig:comb_im}. In the left panel of \fig{fig:comb_im}, we find the EBMs are able to generalize to different sizes at different position (albeit with loss in sample quality) while a conditional model ignores the size latent and generates only images seen in the training. In the right panel of \fig{fig:comb_im}, we found that EBMs are able to generalize to combinations of shape and position by creating a distinctive shape for each conditioned shape latent at different positions (though the generated shape doesn't match the shape of the original shape latent), while a baseline is unable to generate samples. We believe the compositional nature of EBMs is crucial to generalize in this task.
\section{Conclusion}

We have presented a series of techniques to scale up EBM training. We further show unique benefits of implicit generation and EBMs and believe there are many further directions to explore.  Algorithmically, we
think it would be interesting to explore methods for faster sampling, such as adaptive HMC. Empirically, we think it would be interesting to explore, extend, and
 understand results we’ve found, in directions such as compositionality, out-of-distribution detection, adversarial robustness, and online learning. Furthermore, we think it may be interesting to apply EBMs on other domains, such as text and as a means for latent representation learning.

\section{Acknowledgements}

We would like to thank Ilya Sutskever, Alec Radford, Prafulla Dhariwal, Dan Hendrycks, Johannes Otterbach, Rewon Child and everyone at OpenAI for helpful discussions.
\newpage
{\small
\bibliographystyle{plainnat}
\bibliography{reference,ebm}
}


\newpage


\appendix
\section{Appendix}

\subsection{Additional Qualitative Evaluation}
\begin{figure}[h]
\centering
\includegraphics[width=0.75\linewidth]{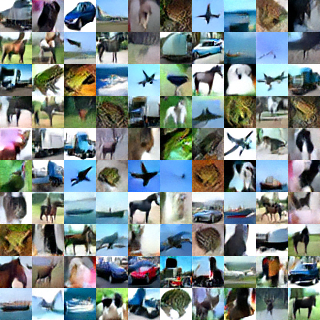}
\caption{\small MCMC samples from conditional CIFAR-10 energy function}
\label{fig:imagenet_uncond}
\end{figure}
\begin{figure}[h]
\centering
\includegraphics[width=0.75\linewidth]{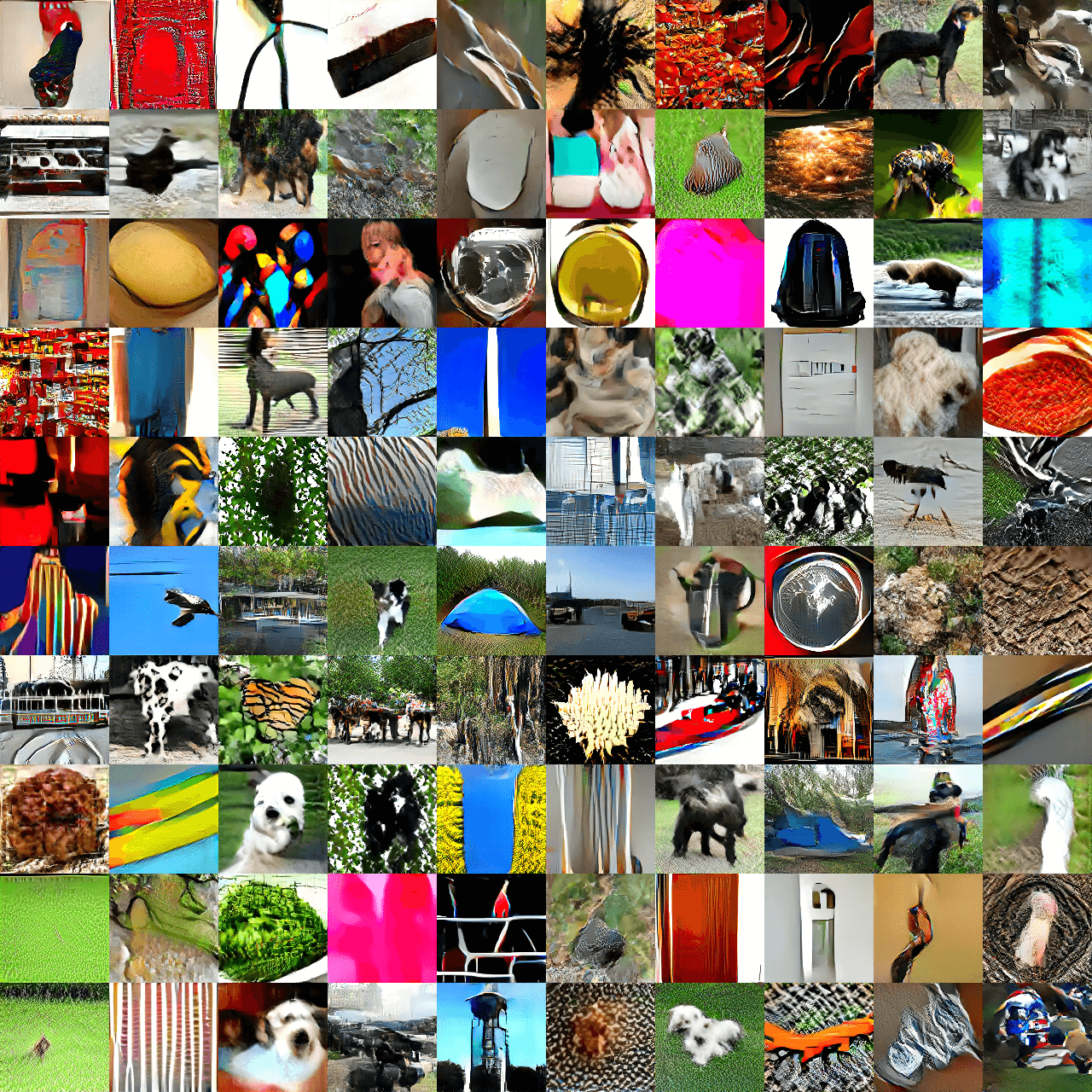}
\caption{\small MCMC samples from conditional ImageNet128x128 models}
\label{fig:imagenet_cond}
\end{figure}
 We present qualitative images from conditional generation on CIFAR10 in \fig{fig:imagenet_uncond} and from conditional generation of ImageNet128x128 in \fig{fig:imagenet_cond}.

\begin{figure}
\begin{center}
\includegraphics[width=0.8\linewidth]{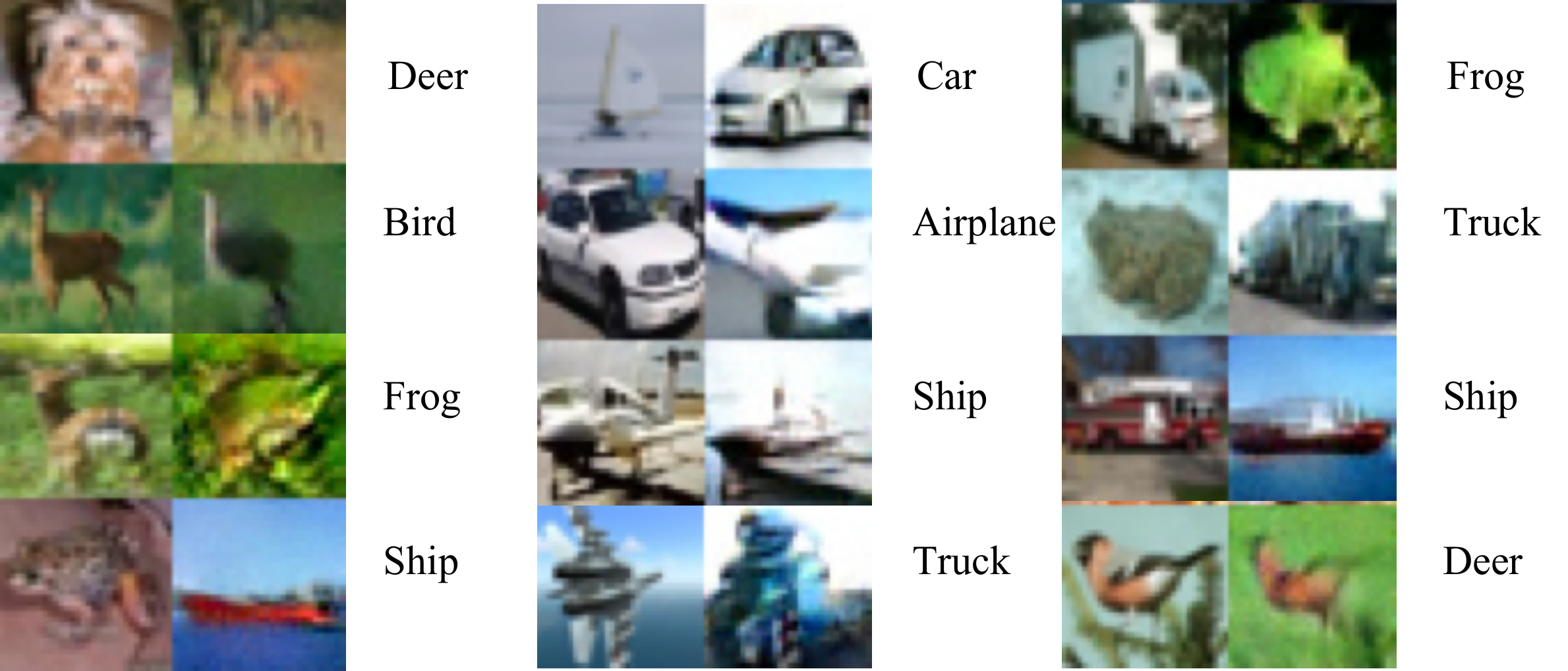}
\caption{\small Illustration of more cross class conversion applying MCMC on a conditional EBM. We condition on a particular class but is initialized with an image from a another class(left). We are able to preserve certain aspects of the image while altering others} 
\label{fig:cross_class_more}
\end{center}
\end{figure}
We provide further images of cross class conversions using a conditional EBM model in \fig{fig:cross_class_more}. Our model is able to convert images from different classes into reasonable looking images of the target class while sometimes preserving attributes of the original class.

\begin{figure*}[h]
\begin{center}
\begin{subfigure}[t]{0.45\textwidth}
    \centering
    \includegraphics[width=\linewidth]{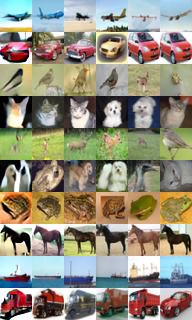}
    \caption{Nearest neighbor images in CIFAR10 for conditional energy models (leftmost generated, seperate class per row).}
\end{subfigure}%
\begin{subfigure}[t]{0.45\textwidth}
    \centering
    \includegraphics[width=\linewidth]{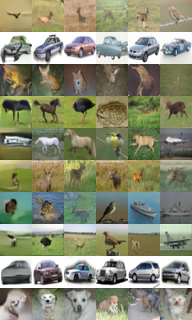}
    \caption{Nearest neighbor images in CIFAR10 for unconditional energy model (leftmost generated)}
\end{subfigure}
\caption{\small Nearest neighbor images $L2$ distance for images generated from implicit sampling.}
\label{fig:nearest_l2}
\end{center}
\end{figure*}

\begin{figure*}[h]
\begin{center}
\begin{subfigure}[t]{0.45\textwidth}
    \centering
    \includegraphics[width=\linewidth]{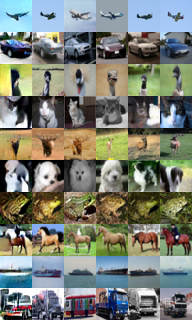}
    \caption{Nearest neighbor images in CIFAR10 for conditional energy models (leftmost generated, seperate class per row).}
\end{subfigure}
\begin{subfigure}[t]{0.45\textwidth}
    \centering
    \includegraphics[width=\linewidth]{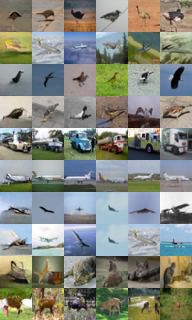}
    \caption{Nearest neighbor images in CIFAR10 for unconditional energy model (leftmost generated)}
\end{subfigure}%
\caption{\small Nearest neighbor images ResNet-50 distance for images generated from implicit sampling.}
\label{fig:nearest_resnet}
\end{center}
\end{figure*}
We analyze nearest neighbors of images we generate in L2 distance \fig{fig:nearest_l2} and in Resnet-50 embedding space in \fig{fig:nearest_resnet}.

\subsection{Test Time Sampling Process}
\begin{figure*}[h]
\begin{center}
\begin{subfigure}[t]{0.5\textwidth}
    \centering
    \includegraphics[width=\linewidth]{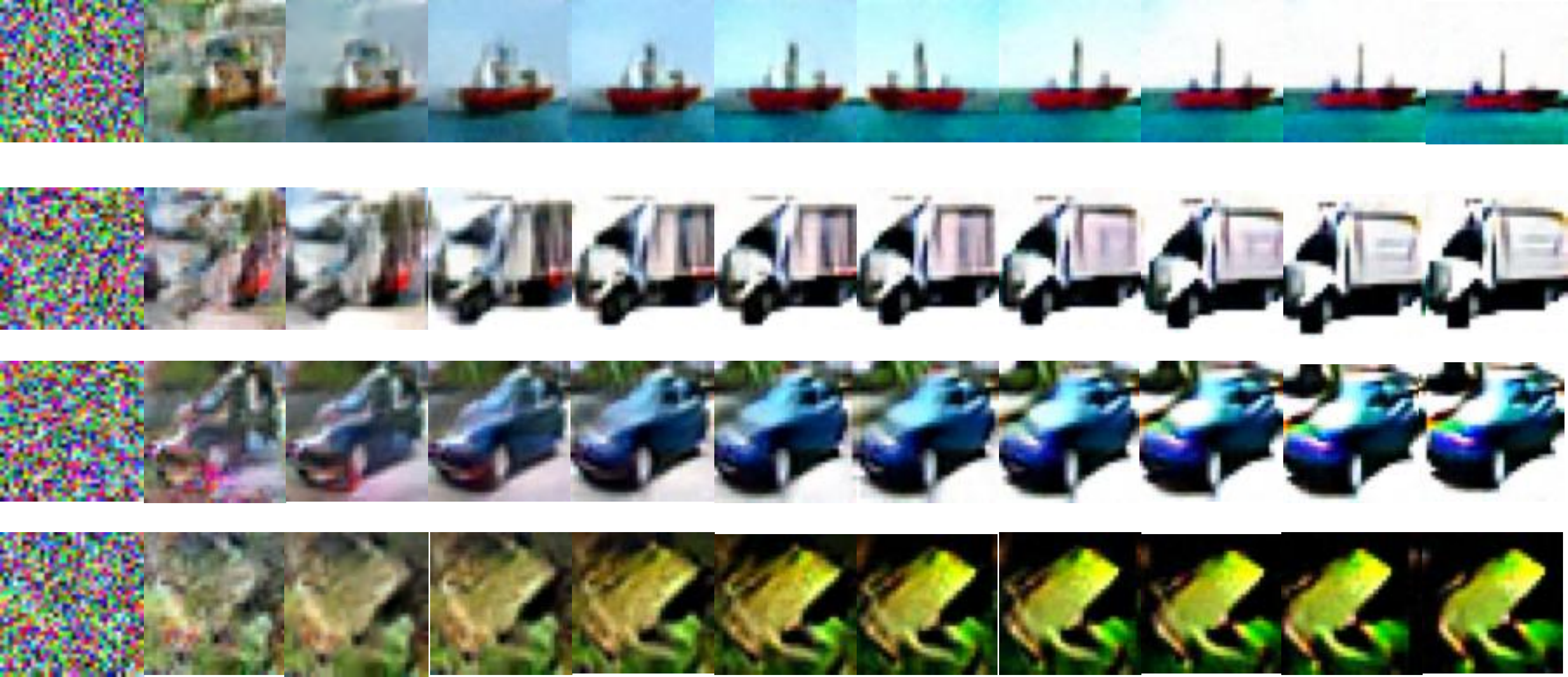}
    \caption{\small Illustration of implicit sampling on conditional EBM of CIFAR-10}
\end{subfigure}%
~
\begin{subfigure}[t]{0.5\textwidth}
    \centering
    \includegraphics[width=\linewidth]{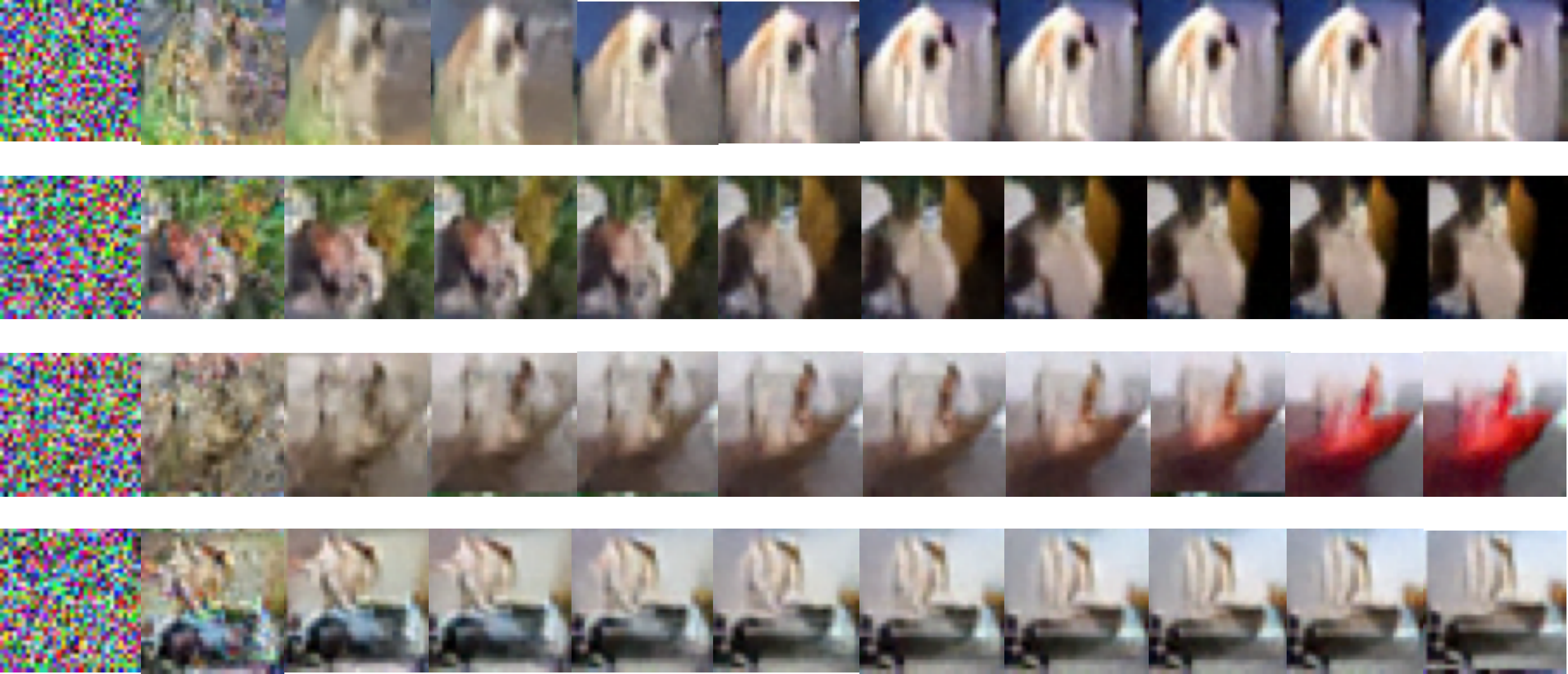}
    \caption{\small Illustration of implicit sampling on an unconditional model on CIFAR-10}
\end{subfigure}
\caption{\small Generation of images from random noise.}
\label{fig:image_gen}
\end{center}
\end{figure*}
We provide illustration of image generation from conditional and unconditional EBM models starting from random noise in \fig{fig:image_gen} with small amounts of random noise added. Dependent on the image generated there is slight drift from some start image to a final generated image. We typically observe that as sampling continues, much of the background is lost and a single central object remains. 

We find that if small amounts of random noise are added, all sampling procedures generate a large initial set of diverse, reduced sample quality images before converging into a small set of high probability/quality image modes that are modes of images in CIFAR10. However, we find that if sufficient noise is added during sampling, we are able to slowly cycle between different images with larger diversity between images (indicating successful distribution sampling) but with reduced sample quality. 

Due to this tradeoff, we use a replay buffer to sample images at test time, with slightly high noise then used during training time. For conditional energy models, to increase sample diversity, during initial image generation, we flip labels of images early on in sampling.

\subsection{Likelihood Evaluation And Ablations}

\begin{figure}
\centering
\begin{subtable}{0.45\textwidth}
    \begin{tabular}{lcc}
    \toprule
    Model & Lower Bound & Upper Bound\\
    \midrule
    EBM + PCD & 380.9 & 482  \\
    GAN 50 \citep{wu2016quantitative} & 618.4 & 636.1 \\
    VAE 50 \citep{wu2016quantitative} & 985.0 & 997.1 \\
    NICE \citep{dinh2014nice} & \textbf{1980.0} & 1980.0 \\
    EBM + Replay Buffer & 1925.0  & \textbf{2218.3} \\
    \bottomrule
    \end{tabular}
\end{subtable}
\caption{\small Log likelihood in Nats on Continuous MNIST. EBMs are evaluated by running AIS for 10000 chains}
\label{tbl:mnist_nll}
\end{figure}

To evaluate the likelihood of EBMs, we use AIS \citep{Neal2001Annealed} and RAISE to obtain a lower bound of partition function \citep{burda2015accurate}. We found that our energy landspaces were smooth and gave sensible likelihood estimates across a range of temperatures and so chose the appropriate temperature that maximized the likelihood of the model. When using these methods to estimate the partition function on CIFAR-10 or ImageNet, we found that it was too slow to get any meaningfull partition function estimates. Specifically, we ran AIS for over 300,000 chains (which took over 2 days of time) and still a very large gap between lower and upper partition function estimates. 

While it was difficult to apply on CIFAR-10, we were able to get lower differences between upper and lower partition functions estimates on continuous MNIST. We rescaled MNIST and to be between 0 and 1 and added 1/256 random noise following \citep{uria2013rnade}. \tbl{tbl:mnist_nll} provides a table of log likelihoods on continuous MNIST across Flow, GAN, and VAE models as well as well as a comparison towards using PCD as opposed to a replay buffer to train on continuous MNIST. We find that the replay buffer is essential to good generation and likelihood,  with the ablation of training with PCD instead of replay buffer getting significantly worse likelihood. We further find that EBMs appear to compare favorably to other likelihood models.

\subsection{Hyper-parameter Sensitivity}

\begin{wrapfigure}{r}{0.5\textwidth}
\vspace{-5mm}
\centering
\resizebox{0.5\textwidth}{!}{
\begin{tabular}{lccc}
    \toprule
     Models & Parameters & Training Time & Sampling Time\\
    \midrule
    EBM & 5M & 48 & 3 Hour (Variable)\\
    PixelCNN++ & 160M & 1300 & 72 Hour \\
    Glow & 115M & 1300 &  0.5 Hour\\
    SNGAN & 5M  & 9 &  0.02 Hour \\
    \bottomrule
\end{tabular}
}
\vspace{-2mm}
\caption{\small Comparison of parameters, training time (GPU hours), and sampling time (for 50000 images) on CIFAR-10. For EBM, sampling time depends on steps of sampling. We used 3 hours of sampling to generate quantitative metrics, but sampling can be much faster (around 0.2 hour) with reduced diversity.}
\label{tbl:model_comparison}
\vspace{-10pt}
\end{wrapfigure}

Empirically, we found that EBM training under our technique was relatively insensitive to the hyper-parameters. For example, \tbl{tbl:hyper} shows log likelihoods on continuous MNIST across several different order of magnitudes of L2 regularization and step size magnitude. We find consistent likelihood and good qualitative generation across different variations of L2 coefficient and step size magnitude and observed similar results in CIFAR-10 and Imagenet. Training is insensitive to replay buffer size (as long as size is greater than around 10000 samples).
\begin{figure}
\centering
\begin{subtable}{0.45\textwidth}
    \begin{tabular}{lccc}
    \toprule
    Hyper-parameter & Value & Lower Bound & Upper Bound\\
    \midrule
    \multirow{3}{*}{L2 Coefficient} & 0.01 & 1519 & 2370 \\
    & 0.1 & 1925 & 2218 \\   
    & 1.0 & 1498 & 2044 \\  
    \midrule
    \multirow{3}{*}{Step Size} & 10.0 & 1498 & 2044 \\
    & 100.0 & 1765 & 2309 \\   
    & 1000.0 & 1740 & 2009 \\  
    \bottomrule
    \end{tabular}
\end{subtable}
\caption{\small Log likelihood in Nats on Continuous MNIST under different settings of the L2 penalty coefficient and Langevin Step Size evaluated after running AIS and RAISE for 10000 chains. Lower and upper bound in likelihood remain relatively constant across several different order of magnitude of variation}
\label{tbl:hyper}
\end{figure}

\subsection{Comparison With Other Likelihood Models}

We  compare EBMs to other generative models in \fig{tbl:model_comparison} on CIFAR-10.  EBMs are faster to train than other likelihood models, with fewer parameters, but are more expensive than GAN based models (due to Langevin dynamics sampling), and slower to sample. Training time for PixelCNN++ and Glow are from reported values in their papers, while sampling time and parameters were obtained from released code repositories. We have added the table to the appendix of the paper and added discussion on these trade-offs and intractability of likelihood evaluation in the main paper.

\subsection{Image Saturation}

\begin{figure}[H]
\vspace{-5mm}
  \begin{center}
    \includegraphics[width=0.5\linewidth]{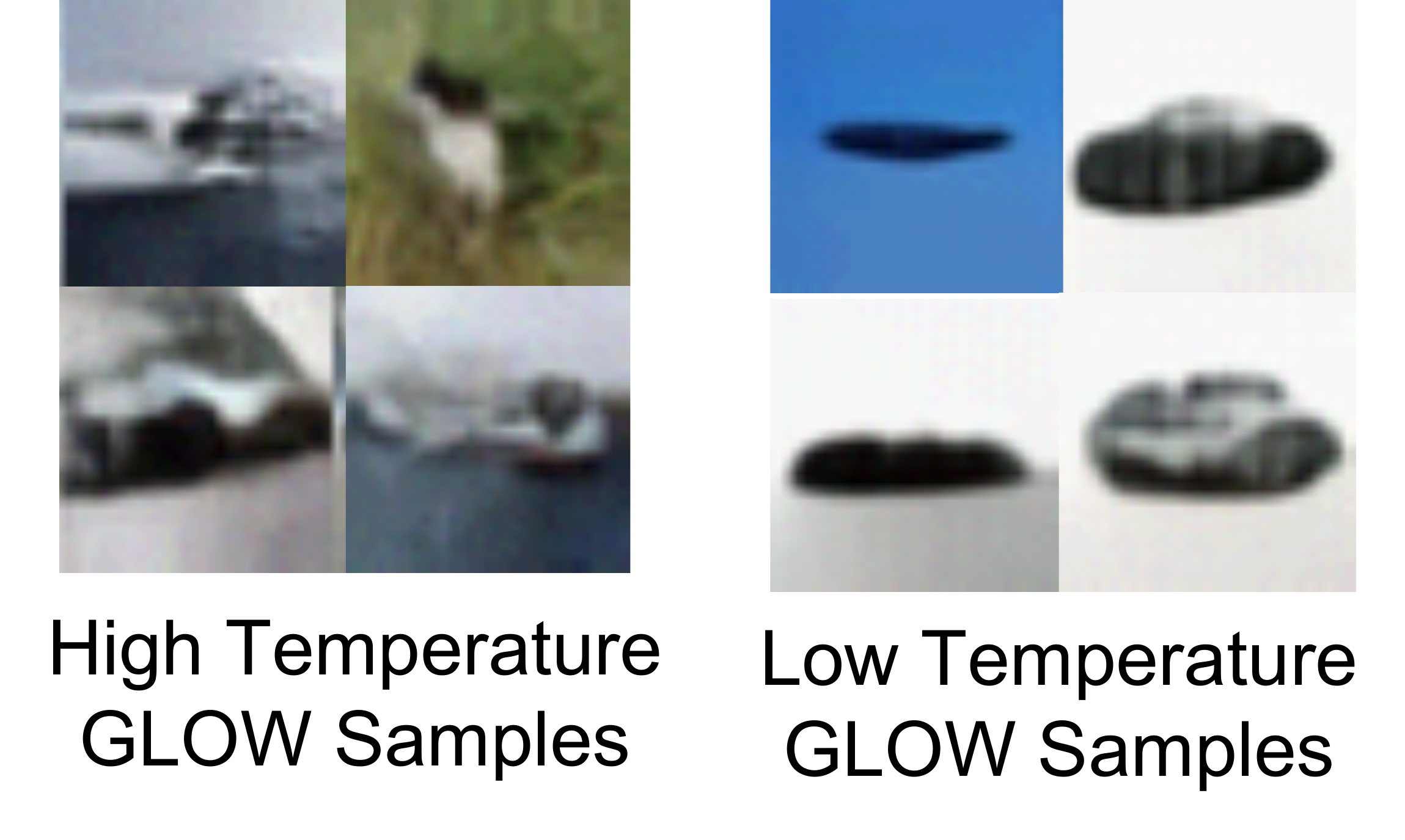}
  \end{center}
  \vspace{-3mm}
  \caption{\small Low Temperature (High likelihood mode) vs High Temperature (Low Likelihood mode) in Glow Model}
  \vspace{-5mm}
  \label{fig:image}
\end{figure}

When EBMs are run for a large number of sampling steps, images appear in increased saturation. This phenomenon can be exampled by the fact that many steps of sampling typically converge to  high likelihood modes. Somewhat unintuitively, as seen also by out of distribution performance of likelihood models, such high likelihood modes on likelihood models trained on real datasets are often very texture based and heavily saturated. We provide illustration of this phenomenon on Glow in \fig{fig:image}.

\subsection{Details on Continual Learning Training}
To train EBM models on the continual learning scenario of Split MNIST, we train an EBM following Algorithm 1 in the main body of the paper. Initially, negative sampling is done with labels of the digits 0 and 1. Afterwards, negative sampling is done with labels of the digits 2 and 3 and so forth. Simultaneously, we train EBMs on ground truth image label annotations. We maintain a replay buffer of negative samples to enable effective training of the EBM.

\subsection{KL Term}

In cases of very highly peaked data, we can further regularize $E$ such that $q$ matches $p$ by minimizing KL divergence between the two distributions:
\begin{align}
\mathcal{L}_\text{KL}(\theta) = \KL{q_\theta}{p} = \E{\sx \sim q_\theta}{\bar{E}(\sx)} + \ent{q_\theta}
\label{eq:kl}
\end{align}
Where $\bar{E}$ is treated as a constant target function that does not depend on $\theta$. Optimizing the above loss requires differentiating through the Langevin dynamics sampling procedure of (\ref{eq:langevin}), which is possible since the procedure is differentiable. Intuitively, we train energy function such that a limited number of gradient-based sampling steps takes samples to regions of low energy. We only use the above term when fine-tuning combinations of energy functions in zero shot combination and thus ignore the entropy term.

The computation of the entropy term $\ent{q_\theta}$ can resolved by approaches \citep{liu2017stein} propose an optimization procedure where this term is minimized by construction, but rely on a kernel function $\kappa(\x,\x’)$, which requires domain-specific design. Otherwise, the entropy can also be resolved by adding a IAF \citep{kingma2016improved} to map to underlying Gaussian through which entropy can be evaluated.

\subsection{Additional Compositionality Results}

\begin{figure}
\begin{center}
\includegraphics[width=1.0\linewidth]{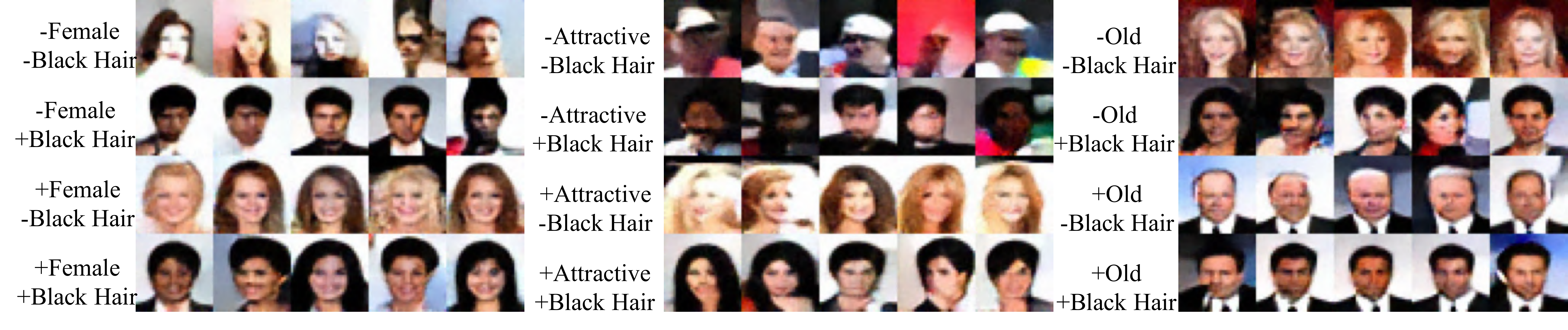}
\caption{\small Illustration of test time generation of various combinations of two independently trained EBMs conditioned on latents on gender, hair color, attractiveness, and age on CelebA.} 
\label{fig:celeba_cond}
\end{center}
\end{figure}

We show additional compositionality results on the CelebA dataset. We trained seperate conditional EBMs on the latents attractiveness, hair color, age, and gender. We show different combinations of two conditional models in \fig{fig:celeba_cond}.

\subsection{Model}
\label{sec:modelarch}
We use the residual model in \fig{fig:cifar10net} for conditional CIFAR-10 images generation and the residual model in \fig{fig:cifar10netnone} for unconditional CIFAR10 and Imagenet images. We found unconditional models need additional capacity. Our conditional and unconditional architectures are  similar to architectures in \citep{miyato2018spectral}. 

\begin{figure*}
\begin{subfigure}[t]{0.25\textwidth}
\centering
\begin{tabular}{c}
    \toprule
    \toprule
    3x3 conv2d, 128 \\
    \midrule
    ResBlock down 128 \\
    \midrule
    ResBlock 128 \\
    \midrule
    ResBlock down 256 \\
    \midrule
    ResBlock 256 \\ 
    \midrule
    ResBlock down 256 \\
    \midrule
    ResBlock 256 \\
    \midrule
    Global Sum Pooling \\ 
    \midrule
    dense $\rightarrow$ 1\\
    \bottomrule
\end{tabular}
\caption{Conditional CIFAR-10 Model}
\label{fig:cifar10net}
\end{subfigure}%
\begin{subfigure}[t]{0.25\textwidth}
\centering
\begin{tabular}{c}
    \toprule
    \toprule
    3x3 conv2d, 128 \\
    \midrule
    ResBlock down 128 \\
    \midrule
    ResBlock 128 \\
    \midrule
    ResBlock 128 \\
    \midrule
    ResBlock down 256 \\
    \midrule
    ResBlock 256 \\ 
    \midrule
    ResBlock 256 \\ 
    \midrule
    ResBlock down 256 \\
    \midrule
    ResBlock 256 \\
    \midrule
    ResBlock 256 \\
    \midrule
    Global Sum Pooling \\ 
    \midrule
    dense $\rightarrow$ 1\\
    \bottomrule
\end{tabular}
\caption{Unconditional CIFAR-10 Model}
\label{fig:cifar10netnone}
\end{subfigure}%
\begin{subfigure}[t]{0.25\textwidth}
\begin{tabular}{c}
    \toprule
    \toprule
    3x3 conv2d, 128 \\
    \midrule
    ResBlock down 256 \\
    \midrule
    ResBlock 256 \\
    \midrule
    ResBlock down 512 \\
    \midrule
    ResBlock 512 \\ 
    \midrule
    ResBlock down 1024 \\
    \midrule
    ResBlock 1024 \\
    \midrule
    Global Sum Pooling \\ 
    \midrule
    dense $\rightarrow$ 1\\
    \bottomrule
\end{tabular}
\caption{Conditional ImageNet32x32 Model}
\label{fig:imagenet_32model}
\end{subfigure}%
\begin{subfigure}[t]{0.25\textwidth}
\begin{tabular}{c}
    \toprule
    \toprule
    3x3 conv2d, 64 \\
    \midrule
    ResBlock down 64 \\
    \midrule
    ResBlock down 128 \\
    \midrule
    ResBlock down 256 \\
    \midrule
    ResBlock down 512 \\ 
    \midrule
    ResBlock down 1024 \\
    \midrule
    ResBlock 1024 \\
    \midrule
    Global Sum Pooling \\ 
    \midrule
    dense $\rightarrow$ 1\\
    \bottomrule
\end{tabular}
\caption{Conditional ImageNet128x128 Model}
\label{fig:imagenet_128model}
\end{subfigure}
\label{fig:architecture}
\end{figure*}

We found definite gains with additional residual blocks and wider number of filters per block. Following \citep{anonymous2019the, kingma2018glow}, we initialize the second convolution of residual block to zero and a scalar multiplier and bias at each layer. We apply spectral normalization on all weights.  When using spectral normalization, zero weight initialized convolution filters were instead initialized from random normals with standard deviations of $1^{-10}$ (with spectrum normalized to be below 1). We use conditional bias and gains in each residual layer for a conditional model. We found it important when down-sampling to do average pooling as opposed to strided convolutions. We use leaky ReLUs throughout the architecture.

We use the architecture in \fig{fig:imagenet_cond} for generation of conditional ImageNet32x32 images.

\subsection{Training Details and Hyperparameters}
\label{sec:hyper}
 For CIFAR-10 experiments, we use 60 steps of Langevin dynamics to generate negative samples. We use a replay buffer of size of 10000 image. We scale images to be between 0 and 1. We clip gradients to have individual value  magnitude of  less than 0.01 and use a step size of 10 for each gradient step of Langevin dynamics. The L2 loss coefficient is set to 1. We use random noise with standard deviation $\lambda = 0.005$. CIFAR-10 models are trained on 1 GPU for 2 days. We use the Adam Optimizer with $\beta_1=0.0$ and $\beta_2=0.999$ with a training learning rate of $10^{-4}$. We use a batch size during training of 128 positive and negative samples. For both experiments, we clip all training gradients that are more than 3 standard deviations from the 2nd order Adam parameters. We use spectral normalization on networks. For ImageNet32x32 images, we an analogous setup with models are trained for 5 days using 32 GPUs. For ImageNet 128x128, we use a step size 100 and train for 7 days using 32 GPUs.
 
 For robotic simulation experiments we used 10 steps of Langevin dynamics to generate negative samples, but otherwise use identical settings as for image experiments.

\subsection{Tips And Failures}

We provide a list of tips, observations and failures that we observe when trying to train energy based models. We found evidence that suggest the following observations, though in no way are we certain that these observations are correct.

We found the following tips useful for training.
\begin{itemize}
    \item We found that EBM training is most sensitive to MCMC transition step sizes (though there is around 2 to 3 order of magnitude that MCMC transition steps can vary). 
    \item We found that that using either ReLU, LeakyReLU, or Swish activation in EBMs lead to good performance. The Swish activation in particular adds a noticeable boost to training stability.
    \item When using residual networks, we found that performance can be improved by using 2D average pooling as opposed to transposed convolutions 
    \item We found that group, layer, batch, pixel or other types of normalization appeared to significantly hurt sampling, likely due to making MCMC steps dependent on surrounding data points.
    \item During a typical training run, we keep training until the sampler is unable to generate effective samples (when energies of proposal samples are much larger than energies of data points from the training data-set). Therefore, to extend training, the number of sampling steps to generate a negative sample can be increased.
    \item We find a direct relationship between depth / width and sample quality. More model depth or width can easily increase generation quality.
    \item When tuning noise when using Langevin dynamics, we found that very low levels of noise led to poor results. High levels of noise allowed large amounts of mode exploration initially but quickly led to early collapse of training due to failure of the sampler (failure to explore modes). We recommend keeping noise fixed at 0.005 and tune the step size per problem (though we found step sizes of around 10-100 work well).
\end{itemize}

We also tried the approaches below with the relatively little success. 

\begin{itemize}
    \item We found that training ensembles of energy functions (sampling and evaluating on ensembles) to help a bit, but was not worth the added complexity.
    \item We found it difficult to apply vanilla HMC to EBM training as optimal step sizes and leapfrog simulation numbers differed greatly during training, though applying adaptive HMC would be an interesting extension.
    \item We didn't find much success with adding a gradient penalty term as it seems to hurt model capacity.
    \item We tried training a separate network to help parameterize  MCMC sampling but found that this made training unstable. However, we did find that using some part of the original model to parameterize MCMC (such as using the magnitude to energy to control step size) to help performance.
\end{itemize}

\subsection{Relative Energy Visualization}

\begin{figure}
\centering
\includegraphics[width=0.6\linewidth]{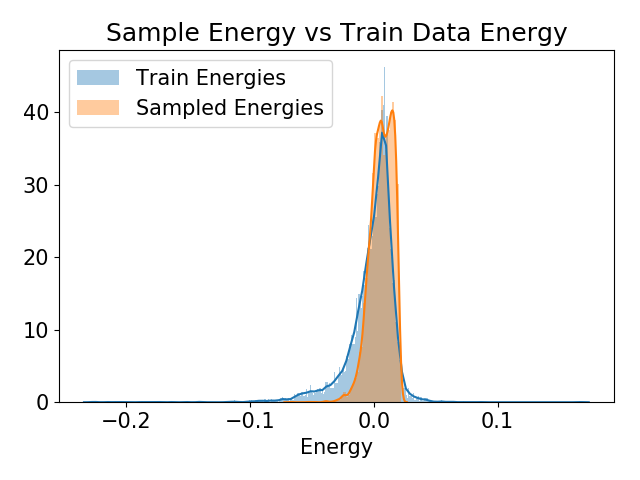}
\vspace{-10pt}
\caption{\small Relative energy of points sampled from $q(x)$ compared to CIFAR-10 train data points. We find that $q(x)$ exhibits a similar distribution to $p_d(x)$ and thus is similar to $p(x)$.}
\label{fig:relative_energy}
\end{figure}
In \fig{fig:relative_energy}, we show the energy distribution from $q(x)$ and from $p_d(x)$. We see that both distributions match each other relatively closely, providing evidence that $q(x)$ is close to $p(x)$

\newpage
\end{document}